\documentclass[sn-mathphys-num]{sn-jnl}

\usepackage{graphicx}
\usepackage{multirow}
\usepackage{amsthm}%
\usepackage{mathrsfs}%
\usepackage{titlesec}
\usepackage[title]{appendix}%
\usepackage{xcolor}%
\usepackage{textcomp}%
\usepackage{longtable}
\usepackage{subcaption}
\usepackage{manyfoot}%
\usepackage{booktabs}%
\usepackage{adjustbox}
\usepackage{algorithm}%
\usepackage{algorithmicx}%
\usepackage{algpseudocode}%
\usepackage{listings}%

\usepackage[utf8]{inputenc}
\usepackage[T5]{fontenc}

\usepackage{longtable}
\usepackage{tabularx} 
\usepackage{enumitem}
\usepackage{hyperref}
\usepackage{amsmath}
\usepackage{adjustbox}
\usepackage{amssymb}
\usepackage{lipsum}
\usepackage{makecell}
\usepackage{stackengine}
\usepackage{lscape}
\usepackage[vietnamese, english]{babel}

\theoremstyle{thmstyleone}%
\theoremstyle{thmstyletwo}%

\theoremstyle{thmstylethree}%

\begin{document}

\title[Article Title]{ViOCRVQA: Novel Benchmark Dataset and Vision Reader for Visual Question Answering by Understanding Vietnamese Text in Images}
\author[1,2]{\fnm{Huy} \sur{Quang Pham}}\email{21522163@gm.uit.edu.vn}

\author[1,2]{\fnm{Thang} \sur{Kien-Bao Nguyen}}\email{21521432@gm.uit.edu.vn}

\author[1,2]{\fnm{Quan} \sur{Van Nguyen}}\email{21521333@gm.uit.edu.vn}

\author[1,2]{\fnm{Dan} \sur{Quang Tran}}\email{21521917@gm.uit.edu.vn}

\author[1,2]{\fnm{Nghia} \sur{Hieu Nguyen}}\email{nghiangh@uit.edu.vn}

\author*[1,2]{\fnm{Kiet} \sur{Van Nguyen}}\email{kietnv@uit.edu.vn}

\author[1,2]{\fnm{Ngan} \sur{Luu-Thuy Nguyen}}\email{ngannlt@uit.edu.vn}

\affil[1]{\orgdiv{Faculty of Information Science and Engineering}, \orgname{University of Information Technology, Ho Chi Minh City, Vietnam}}

\affil[2]{\state{Vietnam National University}, \country{Ho Chi Minh City, Vietnam}}


\abstract{Optical Character Recognition - Visual Question Answering (OCR-VQA) is the task of answering text information contained in images that have just been significantly developed in the English language in recent years. However, there are limited studies of this task in low-resource languages such as Vietnamese. To this end, we introduce a novel dataset, \textbf{ViOCRVQA} (\textbf{Vi}etnamese \textbf{O}ptical \textbf{C}haracter \textbf{R}ecognition - \textbf{V}isual \textbf{Q}uestion \textbf{A}nswering dataset), consisting of \textbf{28,000+} images and \textbf{120,000+} question-answer pairs. In this dataset, all the images contain text and questions about the information relevant to the text in the images. We deploy ideas from state-of-the-art methods proposed for English to conduct experiments on our dataset, revealing the challenges and difficulties inherent in a Vietnamese dataset. Furthermore, we introduce a novel approach, called \textbf{VisionReader}, which achieved 0.4116 in EM and 0.6990 in the F1-score on the test set. Through the results, we found that the OCR system plays a very important role in VQA models on the ViOCRVQA dataset. In addition, the objects in the image also play a role in improving model performance. We open access to our dataset at \href{https://github.com/qhnhynmm/ViOCRVQA.git}{link} for further research in OCR-VQA task in Vietnamese.}



\renewcommand\refname{Abtract}

\maketitle

\section{Introduction}
In recent years, substantial advancements in technology have significantly boosted the productivity of machines, especially in Artificial Intelligence (AI). The elegant combination of Natural Language Processing (NLP) and Computer Vision (CV) has created innovative solutions for many fields. Researchers are increasingly concentrating on developing multimodal models, expanding the ability to understand and respond to questions related to both images and language. This task is not only important in the field of research but also widely applied in everyday life as it resembles the characteristic of human learning: jointly learning from various modalities of information.

In multimodal learning, studies of VQA in English have grown exponentially over the last five years \cite{antol2015vqa, goyal2017making,singh2019towards, biten2019scene} while there is a limited number of works for VQA in low-resource languages \cite{kamel2023vaqa,kim2024bok,
shimizu-etal-2018-visual}, especially in Vietnamese. Despite several recent studies in Vietnamese, its potential remains immense, given that Vietnamese is a rich language that conveys textual meaning. First work on VQA in Vietnamese research starts with the publication of the ViVQA dataset \cite{tran2021vivqa}, then gradually novel datasets were constructed and published such as OpenViVQA \cite{nguyen2023openvivqa}, UIT-EVJVQA \cite{nguyen2023evjvqa}, and ViCLEVR \cite{tran2023viclevr}. One of the latest studies in Vietnamese vision language research is the work of \citet{nguyen2023openvivqa}, which has identified a new task for the VQA task, known as open-ended VQA, in which answers are in open-ended form. With this new form of VQA task, previous methods can not perform effectively on the OpenViVQA dataset \cite{nguyen2023openvivqa}. Moreover, \citet{nguyen2023openvivqa} requires deep learning methods that have the ability to integrate scene texts in images along with objects to give comprehensive answers. Such a task is challenged by the complicated open-ended form of answers as well as the potential prunes of using an external Optical Character Recognition (OCR) system for scene text detection and recognition. Recognizing the unique challenges posed by the OCR-VQA task, we observed that no Vietnamese dataset is currently robust enough to address these issues effectively.

We constructed a novel dataset in the Vietnamese language and called \textbf{ViOCRVQA} (\textbf{Vi}etnamese \textbf{O}ptical \textbf{C}haracter \textbf{R}ecognition - \textbf{V}isual \textbf{Q}uestion \textbf{A}nswering dataset), which aims at enhancing the ability of solving the OCR-VQA task for Vietnamese. The ViOCRVQA dataset contains \textbf{28,282} images and \textbf{123,781} questions relevant to images with answers. To our best knowledge, this dataset is the largest for studying VQA in Vietnamese. Questions related to title, author, publisher, etc. Moreover, we adopted the semi-automatic question-answer generation process to save human annotation time and enrich the dataset with diverse question patterns.

The ViOCRVQA dataset is a high-quality resource for conducting experiments on the ability of the VQA model to understand textual information in the image through the exploration of recent methods in this field. We also conducted a thorough study of the salient features of the dataset and found that objects have a significant influence on the displayed text content. From these insights, we developed a new approach to the ViOCRVQA dataset, which we call \textbf{VisionReader}, based on combining objects and text.

Our main contributions include the following:
\begin{itemize}
\item Constructed the first high-quality large-scale dataset for OCR-VQA task in Vietnamese, focusing on images containing text, especially book covers.

\item Providing information on how to design the experiments as well as evaluating the results of VQA models using multiple SOTA methods on the ViOCRVQA dataset.

\item Developed a new method superior to SOTA methods that are capable of understanding the relationships between objects and text contained in images.

\item Proved the importance of the OCR system in the OCR-VQA task, and the relationship of the object and text in the image makes VQA models generate answers more accurate.

\end{itemize}

\begin{figure*}[h]
  \centering
  \includegraphics[width=1\textwidth]{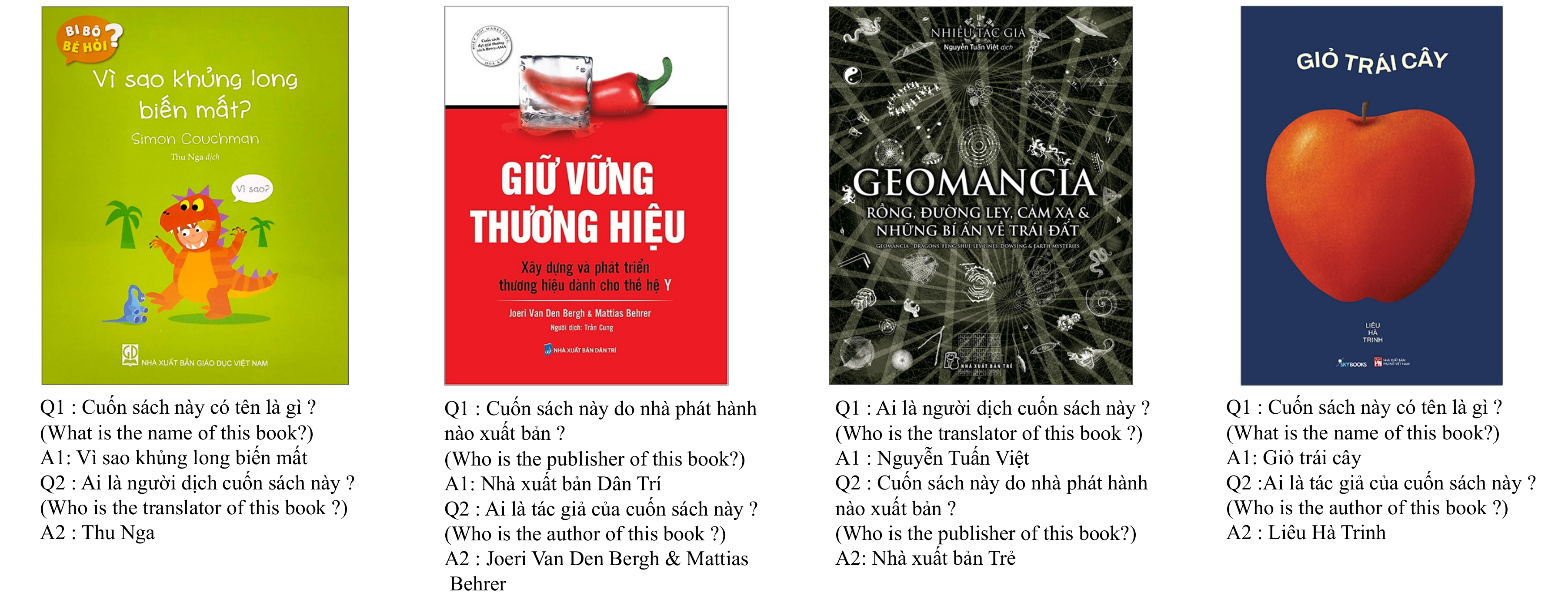}
   \centering\caption{Several examples from the ViOCRVQA dataset.}
  \label{hing}
\end{figure*}

The structure of our article designed as follows:
In Section \ref{related}, we present a literature review of studies in VQA. Detailed descriptions of the ViOCRVQA dataset construction and the method to evaluate the quality of the annotation process are provided in Section \ref{dataset}. Section \ref{metho} introduces our proposed method for evaluating the ViOCRVQA dataset. Section \ref{metric} outlines the evaluation metrics. Section \ref{Experiments and Results} analyzes the main results of the baseline model and our proposed method. Leveraging these results, we deeply analyze the several impacts of VQA models on our dataset in Section \ref{Result Analysis}. Finally, in Section \ref{Conclusion and Future Work}, we suggest future research to address outstanding issues in this dataset.

\section{Related Work}
\label{related}
\subsection{Well-known Visual Question Answering Datasets}
The explosion of the VQA task occurred when \citet{antol2015vqa} released the VQA v1 dataset, with images primarily sourced from the MS COCO \cite{lin2014microsoft} dataset. The introduction of the VQA v1 dataset led to the proposal of numerous methods \cite{kazemi2017show, lu2016hierarchical, kim2016multimodal} for evaluation on this dataset. Later on, \citet{teney2018tips} identified a limitation of the VQA v1 dataset where answers could be transformed into a classification task instead of generating actual answers. To address this flaw, \citet{goyal2017making} introduced the VQA v2 dataset by presenting questions with varied answers across different images, thereby equalizing the frequency of answers for specific types of questions.

Traditional VQA methods have struggled to effectively handle questions that require reading and inference based on text present in images \cite{singh2019towards}. To this end, they introduced the TextVQA dataset, where questions are required to utilize text appearing within the image. To tackle this, they introduced the LoRRA model, which integrates Optical Character Recognition (OCR) to extract and incorporate text from images as answers. Besides that, the ST-VQA \cite{biten2019scene} dataset was created to address the complexity of interpreting and answering questions related to textual content in images. This dataset includes question-answer pairs constructed to require an understanding of the text in images for accurate answers. The goal of this endeavor is to improve model performance in scenarios where understanding scene text is crucial.

In addition, the OCR-VQA-200k dataset \cite{mishra2019ocr} is a focused dataset for the OCR-VQA task in the English language. The dataset is impressive with more than 200,000 images and more than 1 million question-answer pairs. The images of the dataset are mainly book covers, the questions are related to the information on the book covers. However, there are still not many studies related to this dataset.

Not only limited to those VQA datasets, the task has become a hot topic in the global research community, leading to the introduction of numerous other VQA datasets such as DocVQA \cite{mathew2021docvqa}, OpenCQA \cite{kantharaj2022opencqa}, VisualMRC \cite{tanaka2021visualmrc}, InfographicVQA \cite{mathew2022infographicvqa}, providing additional resources for research and the development of methods based on various types of images such as document images, graph images, infographic images, etc.

\subsection{Visual Question Answering Datasets in Vietnamese}
The ViVQA dataset \cite{tran2021vivqa}, marking the early dataset tailored for the VQA task in Vietnamese. The ViVQA dataset is a scaled-down version of the COCO-QA dataset, crafted through semi-automated methods. It comprises 10,328 images and 15,000+ questions, with both the questions and answers characterized by their simplicity.

\citet{nguyen2023evjvqa} introduced multilingual VQA by releasing the EVJVQA dataset, designed for exploring multilingual VQA challenges tailored to the cultural nuances of a specific country. This dataset comprises over 33,000 question-answer pairs in three languages, including Vietnamese, English, and Japanese, which are associated with approximately 5,000 images captured in Vietnam. Notably, the EVJVQA dataset serves as the designated benchmark for the VQA shared task held at the 9th Workshop on Vietnamese Language and Speech Processing.

After that, realizing the limitations of the ViVQA dataset, \citet{nguyen2023openvivqa} introduced the OpenViVQA dataset, the first large-scale handcraft annotation dataset for the VQA task in Vietnamese. This dataset contains over 11,000+ images relevant to more than 37,000+ question-answer pairs. Notably, the answers are not confined to predefined categories, allowing for open-ended responses in various forms of natural language such as words, phrases, or sentences. This departure from answer selection or answer classification in existing VQA datasets adds a new dimension to the challenges and possibilities in the realm of VQA research.

\citet{tran2023viclevr} introduced the ViCLEVR dataset, an
emerging collection for evaluating various visual reasoning capabilities in Vietnamese while mitigating biases. The dataset comprises more than 26,000 images and 30,000 question-answer pairs, each question annotated to specify the type of reasoning involved.

We provide a relative comparison of the OCR-ViVQA dataset with common VQA datasets in English and Vietnamese. Details statistics are listed in Table \ref{dataset_comparisons}.

\begin{table*}[h]
\centering
\setlength{\tabcolsep}{10pt}
\caption{Comparisons VQA datasets in English and Vietnamese.}
\begin{adjustbox}{max width=\columnwidth}
\begin{tabular}{lcrrr}
\hline
\textbf{Dataset}            & \textbf{Language}    & \textbf{Images}             & \textbf{Questions}            & \textbf{Answers}              \\ \hline
VQA v2 \cite{goyal2017making}                      & English              & 204,721                     & 1,105,904                     & 11,059,040                    \\
TextVQA \cite{singh2019towards}                    &                      & 28,408                      & 45,336                        & 453,360                       \\
ST-VQA \cite{biten2019scene}                     &                      & 23,038                      & 31,791                        & 31,791                        \\
DocVQA \cite{mathew2021docvqa}                     &                      & 12,767                      & 50, 000                       & 50, 000                       \\
OCR-VQA-200k \cite{mishra2019ocr}                    &                      & 207,572                     & 1,002,146                     & 1,002,146                     \\
InfographicVQA \cite{mathew2022infographicvqa}             &                      & 5, 485                      & 30,035                        & 30,035                        \\
VisualMRC \cite{tanaka2021visualmrc}                  &                      & 10,197                      & 30,562                        & 30,562                        \\
OpenCQA \cite{kantharaj2022opencqa}                    &                      & 9,285                       &       -                        &           -                    \\
VizWiz \cite{gurari2018vizwiz}                     &                      & 32,842                      & 265,420                       & 265,420                       \\
OK-VQA \cite{marino2019ok}                     &                      & 14,031                      & 14,055                        & 14,055                        \\
GQA \cite{hudson2019gqa}                        &                      & 113,018                     & 22,669,678                    & 22,669,678                    \\
Visual Genome \cite{krishna2017visual}              &  & 108,077 & 1,773,258 & 1,773,258 \\
CLEVR \cite{johnson2017clevr}                      &                      & 100,000                     & 999,968                       & 999,968                       \\ 
\hline
UIT-EVJVQA \cite{nguyen2023evjvqa}                 & Multilingual         & 4,879                       & 33,790                        & 33,790                        \\
MCVQA \cite{hasegawa2023minecraft}                       & &       -  &     369,861     &   369,861       \\ \hline
ViVQA \cite{tran2021vivqa}                      & Vietnamese           & 10,328                      & 15,000                        & 15,000                        \\
OpenViVQA \cite{nguyen2023openvivqa}                  &                      & 11,199                      & 37,914                        & 37,914                        \\
ViCLEVR \cite{tran2023viclevr}                    &                      & 26,000                      & 30,000                        & 30,000                        \\
\textbf{ViOCRVQA (ours)} & \textbf{}            & \textbf{28,282}             & \textbf{123,781}              & \textbf{123,781}              \\ \hline
\end{tabular}
\label{dataset_comparisons}
\end{adjustbox}
\end{table*}

\subsection{Visual Question Answering Methods}
The VQA task, despite numerous advanced methods, remains a challenging task for both the computer vision (CV) and natural language processing (NLP) communities. Given an image and a question in natural language format as input, a VQA model needs to infer the answer based on the image features and linguistic characteristics.

One of the first studies to lay the groundwork for VQA was contributed by \citet{simonyan2015very}, who employed the VGG architecture to extract features from images into smaller image patches. Subsequently, these extracted features are fed into a GRU to process the words in the question, by traversing these images on a per-pixel basis, akin to a snake. Based on this basis, there are a series of other notable studies such as VIS+LSTM \cite{strobelt2017lstmvis}, LSTM Q+I \cite{chen2022quantum}, ABC-CNN \cite{chen2015abc}, Full-CNN \cite{ma2016learning}, LSTM-Att \cite{zhu2016visual7w}, Word + Region \cite{shih2016look}, Attr-CNN+LSTM \cite{wu2017image}, etc. has brought significant progress in solving VQA task.

Since the advent of BERT \cite{kenton2019bert} in encoder transformer style, the continuous development of language models has made the use of traditional RNNs and LSTMs in question processing less common. Their limitations, such as the ability to handle complex contexts and a deep understanding of question meaning, have pushed research on VQA models toward the application of BERT. Subsequent VQA models often use BERT, which has demonstrated a good ability to capture the semantic context of questions. Several notable works in this field include ViLBERT \cite{lu2019vilbert}, VisualBERT \cite{li2019visualbert}, LXMERT \cite{tan2019lxmert}, VL-BERT \cite{su2019vl}, UNITER \cite{chen2020uniter}, OSCAR \cite{li2020oscar}, etc. These models not only solve the problems of traditional models but also open up new potentials in combining visual and textual information more effectively.

Recent studies have shown the effectiveness of applying Transformer language models in encoder-decoder style. As in the case of T5 \cite{raffel2020exploring}, this model yields impressive results which is capable of generating answers when the VQA task is no longer as simple as the classification task before. New research has focused on the application of T5 and incredible progress has been made in this area. Studies such as LaTr \cite{biten2022latr}, PreSTU \cite{kil2023prestu}, VL-T5 \cite{cho2021unifying}, SaL \cite{fang2023separate} are  typical examples of the success of this approach. These studies have greatly contributed to improving the performance of VQA models.

Additionally, when chatGPT was born, it led to an explosion in the race to develop large language models (LLMs) with billions of parameters or more. Studies leveraged LLMs such as BLIP-2 \cite{li2023blip}, Flamingo \cite{alayrac2022flamingo}, Flava \cite{singh2022flava}, mPLUG \cite{li2022mplug}, Palm \cite{maronga2020overview}, Palm 2 \cite{anil2023palm}, etc. which provide incredible performance. However, it should be noted that their implementation requires a large amount of computational resources, which is an important barrier to many researchers.

\section{Dataset Creation}
\label{dataset}
\begin{figure*}[h]
  \centering
  \includegraphics[width=1\textwidth]{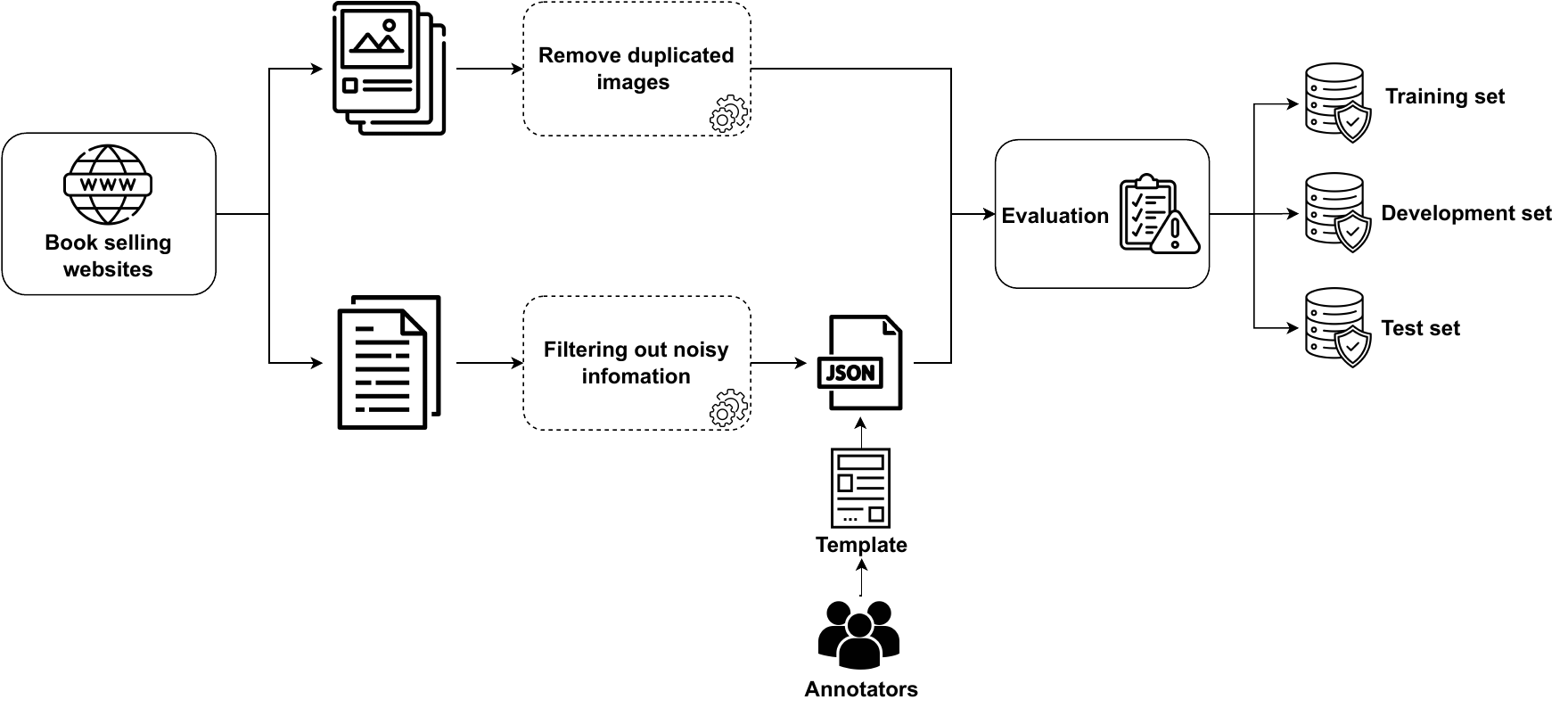}
  \centering\caption{The construction process of the ViOCRVQA dataset.}
  \label{semi-auto}
\end{figure*}

The ViOCRVQA dataset is constructed by a semi-automatic method (Figure~\ref{semi-auto}). In particular, we collected images of book covers from online book-selling websites. On this website, books are displayed with their covers and metadata. We collect their cover images and metadata, then classify information in the metadata into defined categories. Detailed procedures are described as follows.
\subsection{Collecting images and information}
In the OCR-VQA task, the main goal is to focus on extracting information from images containing text, especially from book covers. The reason we choose book covers is because they often contain important information such as title, author, publisher, information about the translator, and more. To collect images, we organized crawling images from online bookstores in Vietnam. We only choose images that contain Vietnamese text to ensure the scope of our research.

\subsection{Data Cleaning}
To initiate the information processing of the books, we commenced by removing punctuation marks and extraneous details that are not presented on the book covers, such as ``tặng kèm'' (``bonus''), ``tái bản 2024'' (``reprinted 2024'') and other irrelevant information. The decision to eliminate punctuation marks was made due to our observation that the information collected contained a significant number of errors in punctuation usage, resulting in data inconsistency compared to the actual information in the book. For example, how the slash is dropped in the title from the metadata differs from how it is on the cover. Throughout the data collection process, we encountered challenges in detecting that the information we gathered contained numerous errors in punctuation usage compared to the actual details in the books. This inconsistency posed challenges to the uniformity of the data, influencing the ability to accurately automate the labeling process. Therefore, the decision to remove all punctuation marks from the dataset is necessary to ensure high quality and consistency in our data processing endeavors.

\subsection{Creating question templates}
We hired ten native Vietnamese speakers, each person was required to annotate at least 30 questions, divided equally into specific fields. These fields include author, book title, publisher, translator, and genres that appear on the book cover. Through the process of careful ideation and compilation, each annotator came up with creative and diverse questions, aiming to make the question set not only rich in content but also attractive to the reader. For example, one of the questions about the author could be ``cuốn sách này do nhà xuất bản nào chịu trách nhiệm phát hành?'' (``which publisher is responsible for publishing this book?'') instead of ``tên nhà xuất bản?'' (``name of publisher?'').

We collected more than 60 unique questions, which are created by annotators for each field, in other word we got a total of 300 rich and diverse questions. Then, we randomly selected these questions and combined them with information from corresponding books on each field.
The questions in our dataset are divided into five categories:
\begin{enumerate}[label=\textbullet]
    \item \textbf{Author:} Questions relate to the author of the book.
    \item \textbf{Title:} Questions relate to the title of the book.
    \item \textbf{Publisher:} Questions relate to the publisher of the book.
    \item \textbf{Translator:} Questions relate to the translator of the book.
    \item \textbf{Genre:} Questions relate to the genre of the book.
\end{enumerate}

\subsection{Statistics}
\begin{table}[htp]
\caption{Statistic information of the ViOCRVQA dataset.}
    \centering{
        \begin{tabular}{lcccl} \hline
                   & \textbf{Train} & \textbf{Dev} & \textbf{Test} & \textbf{Total} \\ \hline
        \textbf{Images}    & 19,798         & 4,243        & 4,241         & 28,282         \\ \hline
        \textbf{Questions} & 86,592         & 18,587       & 18,601        & 123,781        \\
        \textbf{Answers}   & 86,592         & 18,587       & 18,601        & 123,781 \\ \hline      
        \end{tabular}
    }
    \label{statistics}
\end{table}
The ViOCRVQA dataset includes 28,282 images, accompanied by 123,781 questions-answers pairs. About 30\% of the total images and the entire questions, along with their corresponding answers, were selected to form the validation set and test set. Each partition accounts for about 15\% of the total images, and the remaining images are retained for the training set. A random selection of images for the test and validation sets was performed using a uniform distribution. Table \ref{statistics} presents the size of the training, development, and test sets, including the number of images and corresponding question-answer pairs.
\begin{table}[!ht]
\centering
\caption{Distribution of questions and answers among aspects of questions.}
    \centering
    \begin{tabular}{clcc}
        \hline
        \textbf{\#} & \textbf{Aspect} & \textbf{Questions} & \textbf{Answers} \\ \hline
        1           & Author          & 27,881                   & 27,881                 \\
        2           & Title           & 28,283                   & 28,283                \\
        3           & Publisher       & 28,283                   & 28,283                \\
        4           & Translator      & 11,051                   & 11,051                \\
        5           & Genre            & 28,283                   & 28,283                \\ \hline
    \end{tabular}
    \label{distribution_type_ques}
\end{table}

As depicted in Table \ref{distribution_type_ques}, in the Vietnamese context, it clearly shows that the number of questions of the translator is comparatively lower than the other four question types. 

\begin{table}[htp]
\centering
\caption{Composition of ViOCRVQA dataset in brief.}
\begin{tabular}{lr}\hline
Unique author                     & 12,371         \\
Unique title                      & 26,713        \\
Unique publisher                  & 176         \\
Unique translator                 & 3,713          \\
Unique genre                      & 32             \\
Average question length (in words) & 9.64         \\
Average answer length (in words)             & 7.52         \\
Average number of questions per image        & 4.37         \\ \hline
    \end{tabular}
    \label{distribution_1}
\end{table}

In addition, Table ~\ref{distribution_1} shows the diversity and richness of the ViOCRVQA dataset. With 12,371 authors, 26,713 different titles, and nearly 200 publishers. Especially, the number of 3,713 different translators, highlighting the linguistic diversity when converting to Vietnamese. The average length of questions and answers is 9.64 and 7.52, respectively. These are important numbers, hinting at the detail and complexity of the information in the dataset. Each photo in the dataset contains, on average, 4.37 questions and related answers, demonstrating a high level of interaction between images and language.

\subsection{Dataset Comparison}

 \begin{figure*}[htp]
  \centering
  \includegraphics[width=1.0\textwidth]{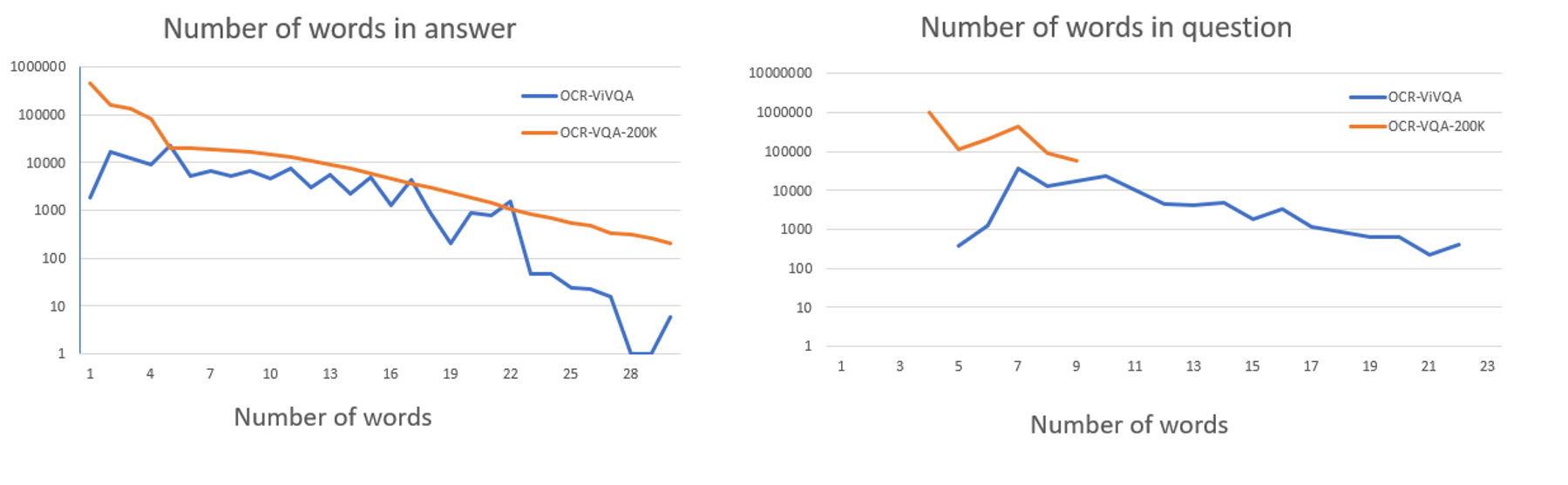}
  \centering \caption{Distribution ViOCRVQA and OCR-VQA-200K}
  \label{Dataset_Comparation}
\end{figure*}

To get an overview of our dataset and another dataset in the same domain in a resource-rich language like English, we compared distribution length with the famous OCR-VQA dataset. Figure \ref{Dataset_Comparation} shows that although the number of images is not too large, it provides a large number of question-answer pairs, demonstrating the effective utilization of information obtained from the images.

Both the OCR-VQA-200K \cite{mishra2019ocr} and ViOCRVQA datasets have different distribution lengths. However, our dataset is superior in terms of question and answer length diversity. This diversity can be attributed to the involvement of 10 Vietnamese annotators who created a total of 300 unique questions, deeply infused with the nuances of Vietnamese linguistic and cultural essence. This effort has endowed the dataset with a richer representation, accurately reflecting the linguistic diversity inherent within the Vietnamese community. 

\subsection{Linguistic Level}
The Linguistic Complexity Specification (LCS) methodology, as presented by \citet{nguyen2023openvivqa}, evaluates sentence complexity by analyzing statistical interactions among tokens and utilizing dependency parsing to frame semantic structures. In table \ref{LCS}, compare the LCS of ViOCRVQA and other VQA datasets.
\begin{table*}[h]
\centering
\caption{Linguistic comparison on questions and answers among VQA datasets. Note that these results were obtained on train-dev sets.}
\resizebox{\textwidth}{!}{
\begin{tabular}{cllcccccccccc}
\hline
\multicolumn{3}{c}{\multirow{2}{*}{Dataset}}  & \multicolumn{1}{c}{\multirow{2}{*}{Language}} & \multicolumn{3}{c}{Word} & \multicolumn{3}{c}{Dependency} & \multicolumn{3}{c}{Height} \\

\multicolumn{3}{c}{}        &\multicolumn{1}{c}{}    & min.   & mean.   & max.  & min.     & mean.     & max.    & min.    & mean.   & max.   \\ \hline
\multicolumn{2}{c}{\multirow{6}{*}{Question}} & VQA v2 \cite{goyal2017making}  & English                & 2      & 6.2     & 23    & 2        & 6.3       & 26      & 1       & 3.3     & 14     \\
\multicolumn{2}{c}{}                          & TextVQA \cite{singh2019towards} & English              & 2      & 7.1     & 33    & 2        & 7.5       & 39      & 1       & 3.9     & 21     \\
\multicolumn{2}{c}{}                          & OCR-VQA \cite{mishra2019ocr} & English              & 4      & 6.5     & 9     & 4        & 6.5       & 10      & 2       & 3.6     & 6      \\
\multicolumn{2}{c}{}                          & ViVQA  \cite{tran2021vivqa}& Vietnamese                & 3      & 9.5     & 24    & 2        & 7.3       & 23      & 2       & 5.5     & 14     \\
\multicolumn{2}{c}{}                          & OpenViVQA \cite{nguyen2023openvivqa} & Vietnamese            & 3      & 10.1    & 32    & 2        & 7.8       & 27      & 2       & 5.2     & 16     \\
\multicolumn{2}{c}{}                          & ViCLEVR \cite{tran2023viclevr}   & Vietnamese          & 3      & 18.57    & 45    & 5        & 16.77       & 40      & 2       & 3.88     & 10    \\
\multicolumn{2}{c}{}                          & \textbf{ViOCRVQA (ours)}& Vietnamese        & \textbf{5}      & \textbf{9.6}     & \textbf{22}    & \textbf{4}        & \textbf{7.3}      & \textbf{17}      & \textbf{1}       & \textbf{1.0}     & \textbf{3}      \\ \hline
\multicolumn{2}{l}{\multirow{6}{*}{Answer}}   & VQA v2  \cite{goyal2017making} & English               & 1      & 1.2     & 18    & 0        & 2.8       & 44      & 1       & 1.0     & 11     \\
\multicolumn{2}{l}{}                          & TextVQA    \cite{singh2019towards}  & English          & 1      & 1.6     & 85    & 0        & 1.5       & 103     & 1       & 1.3     & 40     \\
\multicolumn{2}{l}{}                          & OCR-VQA   \cite{mishra2019ocr} & English             & 1      & 3.3     & 74    & 0        & 2.8       & 100     & 1       & 1.8     & 38     \\
\multicolumn{2}{l}{}                          & ViVQA   \cite{tran2021vivqa}  & Vietnamese             & 1      & 1.8     & 4     & 0        & 0.5       & 3       & 1       & 1.5     & 3      \\
\multicolumn{2}{l}{}                          & OpenViVQA \cite{nguyen2023openvivqa} & Vietnamese            & 1      & 6.9     & 54    & 0        & 4.8       & 52      & 1       & 4.0     & 22     \\
\multicolumn{2}{l}{}                          & ViCLEVR \cite{tran2023viclevr} & Vietnamese            & -      & -    & -    & -        & -       & -      & -       & - & -     \\
\multicolumn{2}{l}{}                          & \textbf{ViOCRVQA (ours)}& Vietnamese        & \textbf{1}      & \textbf{7.5}     &\textbf{55}    & \textbf{0}        & \textbf{4.9}       & \textbf{49}      & \textbf{1}       & \textbf{1.1}     & \textbf{5}     \\ \hline
\end{tabular}%
}
\label{LCS}
\end{table*}

The Linguistic Level Specification (LLS) method \cite{nguyen2023openvivqa} classifies text into categories such as words, phrases, or sentences based on dependency parsing. By implementing LLS, we can discern the prevalent linguistic level of sentences that humans typically choose in answer to questions, underscoring the inherent natural diversity of human answers. 

\begin{table}[!h]
\caption{Linguistic level comparison among VQA datasets. Note that these results were obtained on train-dev sets.}
\label{LLS}
\begin{tabular}{lcrrr}
\hline
\multicolumn{1}{c}{\textbf{Dataset}} & \multicolumn{1}{c}{\textbf{Language}} & \multicolumn{1}{c}{\textbf{word}} & \multicolumn{1}{c}{\textbf{phrase}} & \multicolumn{1}{c}{\textbf{sentence}} \\ \hline
VQA v2\cite{goyal2017making} & English  & 5,884,207 & 651,128 & 45,775 \\
OCR-VQA \cite{mishra2019ocr} & English  & 3,287 & 302,497 & 15,010 \\
Text-VQA\cite{singh2019towards} & English  & 28,317 & 35,964 & 4,947 \\
ViVQA \cite{tran2021vivqa} & Vietnamese  & 3,276 & 6,321 & 0 \\
OpenViVQA \cite{nguyen2023openvivqa} & Vietnamese  & 1,067 & 21,022 & 12,289 \\ \hline
\textbf{ViOCRVQA (ours)} & Vietnamese  & \textbf{2,884} & \textbf{96,612} & \textbf{5,683} \\ \hline
\end{tabular}

\end{table}

Table \ref{LLS} shows that a significant proportion, exceeding 90\%, of the answers in the ViOCRVQA dataset were comprised of phrases. This finding aligns closely with observations from the ViOCRVQA dataset, where a substantial majority of responses also constituted phrases. This prevalence of phrase-centric answers underscores a common trend in our dataset.

Comparatively, the proportion of phrase-centric answers in other datasets is notably lower. For instance, in the Text-VQA dataset, only 55\% of answers are identified as phrases, while in the ViVQA dataset, this figure stood at 66\%. Even in the openViVQA dataset, which answers questions in free-form format, the prevalence of phrase-centric answers was only at 70\%. This difference may stem from the unique characteristics and nuances of each dataset such as question type, answer format, and data collection method. Understanding these differences is critical to developing robust models capable of effectively handling the wide variety of responses in our ViOCRVQA dataset.

\subsection{Human Evaluation}
\begin{table}[htp]
\centering
\caption{Results of human evaluation according to different fields.}
\begin{tabular}{lccc}
\hline
\textbf{Field}    & \textbf{EM (\%)} & \textbf{F1-score (\%)} \\ \hline
Author     & 92.09             & 93.32                  \\
Title      & 88.98             & 92.20                  \\
Publisher  & 91.01             & 93.83                 \\
Translator & 89.09             & 91.70               \\
Type       & 88.48             & 89.29               \\ \hline
\textbf{Average}    & \textbf{89.93}             & \textbf{92.07}              \\ \hline
\end{tabular}
\label{result_human}
\end{table}

For semi-automatic generated VQA datasets, one important aspect to consider is the evaluation of dataset quality, specifically assessing whether the assigned answers truly correspond to the text appearing in the images. To conduct this evaluation, we invited five native Vietnamese speakers from our city. For each type of question, we randomly sampled 100 question-answer pairs from our dataset and then removed the answers. Participants then independently wrote their own answers based on the question content and the text appearing in the images, which were then compared to our automatically annotated answers.

To evaluate the quality of semi-automatic generated answers compared to expert-provided answers, we use F1-score and Exact Match metrics which are mentioned in Section \ref{metric}. After receiving results from the invited annotators, we proceed with the evaluation and obtain detailed results as described in Table \ref{result_human}. The results show that the information we obtained through automatic labeling is consistent with the information manually labeled by humans. This is evidenced by the EM and F1-score reaching very high.

\section{Methodology}
\label{metho}
\begin{figure*}[ht]
  \centering
  \includegraphics[width=1.0\textwidth]{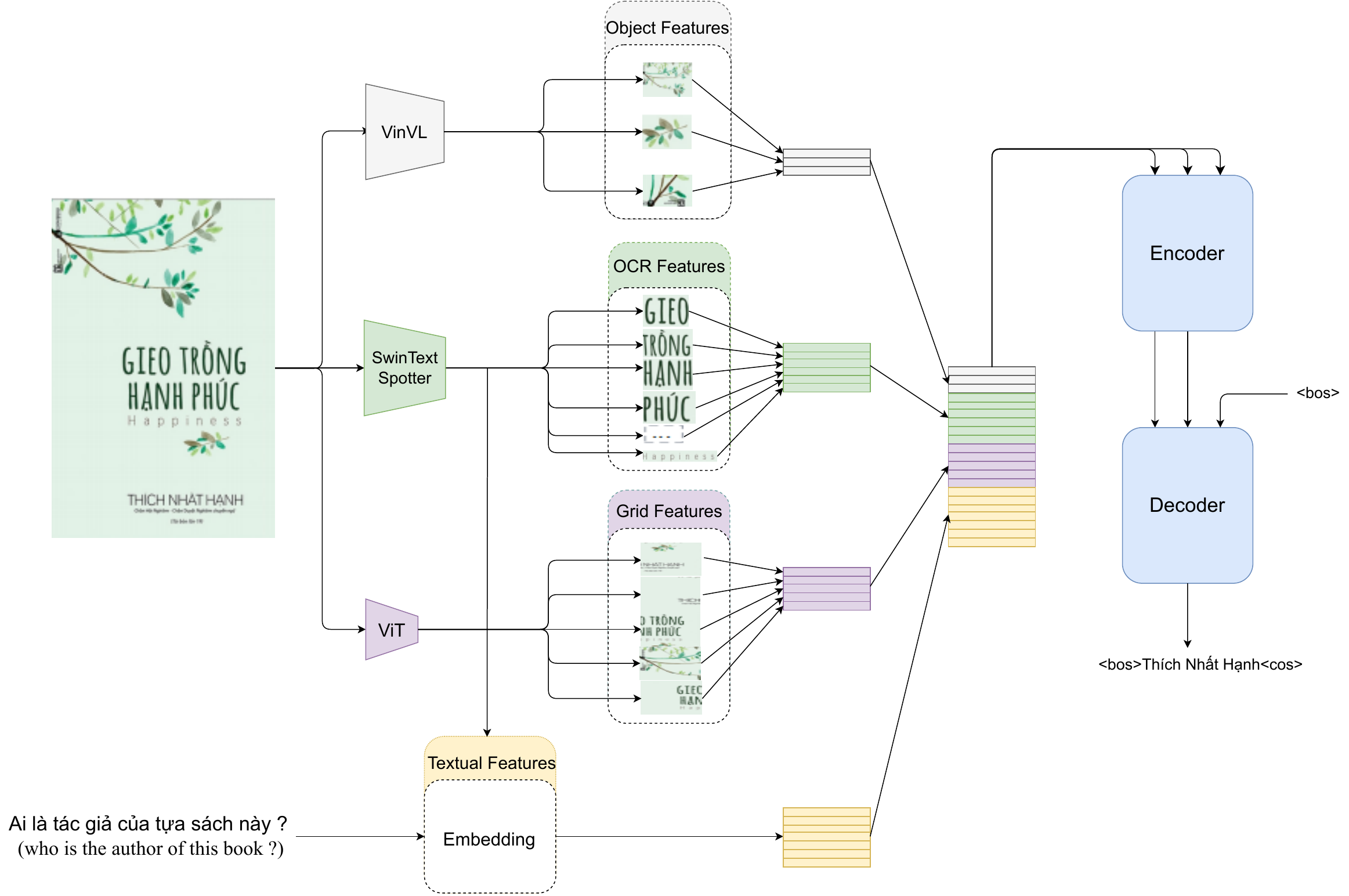}
  \caption{Overview of VisionReader structure.}
  \label{VisionReader structure}
\end{figure*}

We assume that the object plays an important role in determining what information is on the book cover. Therefore, we propose VisionReader centers around the implementation of transformer-based encoder-decoder approach. Following the previous studies in text-based VQA task, we deployed to evaluate methods having \textbf{ViT5} \cite{phan2022vit5} and \textbf{BARTpho} \cite{bartpho} as encoder-decoder module (see Figure \ref{VisionReader structure}). \textbf{ViT5}, based on the \textbf{T5} 
\cite{raffel2020exploring} architecture, and \textbf{BARTPho}, a variant of the \textbf{BART} \cite{lewis2020bart} architecture, have both been trained on a large of Vietnamese text data. These models stand out as the preeminent and potent pre-trained language models for the Vietnamese language, ensuring the efficacy and robustness of our proposed method. 

In our VisionReader, we employed VinVL \cite{zhang2021vinvl} to capture object features and SwinTextSpotter \cite{huang2022swintextspotter} to obtain OCR features. In addition to these, we utilized ViT \cite{dosovitskiy2020image} for the grid features extraction process. For text information, we leveraged the token embedding layer of the language model for processing. The resulting textual features, grid features, object features, and OCR features were concatenated together to form the input for the encoder-decoder module. Integrating object and OCR features in our VisionReader has brought a significant improvement in the ability to understand and process information from questions and images in OCR-VQA task. The results in Section \ref{sec:experiment} show that our proposed method brings significant progress.

\subsection{Multimodal Features Embedding}
\textbf{Object features:} For each given image, the VinVL model processes the image to extract region object features, resulting in a set \(R = \{r_1, r_2, ..., r_k\}\), where each \(r_i\) is a 2048-dimensional vector corresponding to an object in the image. To standardize the bounding box coordinates of the objects, each bounding box \(b_i\) is defined as \(\left[\frac{{x_i^{min}}}{w}, \frac{{y_i^{min}}}{h}, \frac{{x_i^{max}}}{w}, \frac{{y_i^{max}}}{h}\right]\), with \(x_i\) and \(y_i\) denoting the coordinates, and \(w\) and \(h\) representing the width and height of the image, respectively. The set of normalized object bounding boxes is denoted as \(B_{\text{obj}} = \{b_1, b_2, ..., b_k\}\), where each \(b_i\) is a 4-dimensional vector.

The final object features, \(V_{\text{obj}}\), are computed by combining the region object features (\(R'\)) with their respective normalized bounding boxes (\(B'_{\text{obj}}\)). Here, \(B'_{\text{obj}}\) and \(R'\) are obtained by applying a linear layer to \(B_{\text{obj}}\) to project it to the dimension of the language model in \(\mathbb{R}^d\). Therefore, this formula as:

\begin{equation}
V_{\text{obj}} = R' + B'_{\text{obj}}
\end{equation}

\textbf{OCR features:} We employed the SwinTextSpotter model, designed to proficiently handle the Vietnamese language in optical character recognition (OCR) task. Employing this model on each image, we acquired detection features \(D = \{d_1, d_2, ..., d_m\}\) and recognition features \(Z = \{z_1, z_2, ..., z_m\}\), where \(d_i\) and \(z_i \in \mathbb{R}^{256}\). These features are associated with a collection of OCR texts corresponding to their respective bounding boxes. To maintain consistency with object bounding boxes, we normalized the OCR bounding boxes, denoted as \(B_{\text{ocr}} = \{o_1, o_2, ..., o_m\}\), where each \(o_i\) is a 4-dimensional vector.

The composite OCR features are defined as the concatenation of the normalized detection features and recognition features, each augmented with their corresponding bounding box features. Mathematically, this is expressed as: 

\begin{equation}
 S_{\text{ocr}} = \text{Concat}(D' + B'_{\text{ocr}}, Z' + B'_{\text{ocr}})
\end{equation}

where \(D'\), \(Z'\), and \(B'_{\text{ocr}}\) are obtained by applying linear layers to project them to the dimension of the language model in \(\mathbb{R}^d\).

\textbf{Textual features:} The question and the OCR text are embedded through the token embedding layer of the language model. This process yields textual features denoted as \(T = \{t_1, t_2, \ldots, t_n\}\), where \(t_i \in \mathbb{R}^d\) and \(1 \leq \text{len}(T) \leq L\). In this context, \(t_i \in \mathbb{R}^d\) signifies the embedding of the \(i^{th}\) word in the input text, \(L\) represents the length of the input text, and \(d\) stands for the dimensionality of the language model.

\textbf{Grid features:} We chose to use ViT as an important part of our approach because it offers many important advantages. ViT has been proven effective in image processing as well as grid features extraction. With its attention mechanism, ViT is capable of capturing both global and local information in the image, thereby increasing flexibility in identifying important features and helping to improve the ability to understand the picture of our proposed method. Using ViT by freezing it and projecting the last hidden state vector to the dimension of the language model, we obtain the grid features denoted as \(V\).

Therefore, the input embedding fed into the encoder-decoder module is:

\begin{equation}
\text{Input} = \text{Concat}(T,V,V_{\text{\text{obj}}},S_{\text{ocr}})
\end{equation}

where $T$ is textual features, $V$ is grid features extract by ViT, $ V_{\text{obj}}$ is VinVL region object features, $S_{\text{ocr}}$ is SwinTextSpotter OCR features. The \(\text{Concat}(\cdot)\) stands for the concatenating
function.

\subsection{Encoder-Decoder Module}
In the OCR-VQA task, we employed the transformer encoder-decoder architecture, which is used in ViT5 \cite{phan2022vit5} and BARTPho \cite{bartpho} for the encoder-decoder module of VisionReader. The encoder receives the input features and then passes them to the decoder to generate the output sentence. In the decoder, attention mechanisms are employed, directing focus to both the output of the encoder and the input of the decoder.

\textbf{Multi-Head Attention:}
\begin{equation}
\text{MHAtt}(Q, K, V) = \text{Concat}(\text{head}_1, \text{head}_2, ..., \text{head}_h) \cdot W^O
\end{equation}
\begin{equation*}
\text{where } \text{head}_i = \text{Attention}(Q \cdot W^Q_i, K \cdot W^K_i, V \cdot W^V_i)
\end{equation*}

\textbf{Encoder:}
\begin{equation}
\text{EnOut} = \text{LayerNorm}(\text{MHAtt}(X, X, X) + X)
\end{equation}

\begin{equation*}
\text{EnOut} = \text{LayerNorm}(\text{FeedForward}(\text{EnOut}) + \text{EnOut})
\end{equation*}

\textbf{Decoder:}
\begin{equation}
\text{DeOut} = \text{LayerNorm}(\text{MHAtt}(Y, Y, Y) + Y)
\end{equation}

\begin{equation*}
\text{DeOut} = \text{LayerNorm}(\text{MHAtt}(\text{DeOut}, \text{EnOut}, \text{EnOut}) + \text{DeOut})
\end{equation*}

\begin{equation*}
\text{DeOut} = \text{LayerNorm}(\text{FeedForward}(\text{DeOut}) + \text{DeOut})
\end{equation*}

In these equations, \(Q\), \(K\), and \(V\) denote the query, key, and value matrices respectively. \(W^Q_i\), \(W^K_i\), \(W^V_i\), and \(W^O\) are learnable weight matrices specific to each attention head \(i\). \(X\) represents the input features, \(Y\) represents the target output sequence. The function \(\text{Concat}(\cdot)\) is concatenating
function, and \(\text{Attention}(\cdot)\) computes the attention mechanism. \(\text{LayerNorm}(\cdot)\) represents layer normalization.

\section{Evaluation Metrics}
\label{metric}
On the ViOCRVQA dataset, answers are exactly OCR tokens in images. The VQA methods must give exact answers to the given questions as the task definition of the ViOCRVQA dataset. To this end, we use \textbf{Exact Match} and \textbf{F1-score} to evaluate approaches on our dataset. Note that previous studies of VQA in English only use EM as a metric.

\subsection{Exact Match}
\label{EM}
Exact Match (EM) requires ground-truth answers and the predicted answers must be exactly the same. In particular, let $GA=\{ga_1, ga_2,...,ga_n\}$ the set of all ground truth answers, $PA=\{pa_1,pa_2,...,pa_n\}$ the set of all respective predicted answers, EM is determined by:

\begin{equation}
    EM(ga_i,pa_i) = \frac{1}{n}\sum_{i=1}^n{\delta(ga_i,pa_i)}
\end{equation}

where $n \in \mathbb{N}$ is the total number of answers, $\delta$ is the Kronecker function with $\delta(x,y)=1 \Leftrightarrow x=y$ and $\delta(x,y)=0$ otherwise.

\subsection{F1-score}
\label{F1-score}
F1-score is the harmonic mean of Precision and Recall. On our dataset, we define the Precision and Recall at the token level. Given a sentence, its tokens are determined by splitting it by space.

Let $GA$ and $PA$ defined as above, for any $ga_i \in GA$ and $pa_i \in PA$, $ga_i \cap pa_i$ is the set of mutual tokens whose size is defined as $|ga_i \cap pa_i|$. The Precision and Recall at the token level is defined as:
 
\begin{equation}
    P_i = \frac{|ga_i \cap pa_i|}{|pa_i|}; R_i = \frac{|ga_i \cap pa_i|}{|ga_i|}
\end{equation}

The F1-score at the token level is then defined as:

\begin{equation}
    \text{F1}_i = \frac{P_i \times R_i}{P_i + R_i}
\end{equation}

and the overall F1-score on the dataset is defined as:

\begin{equation}
    \text{F1-score} = \frac{1}{n}\sum_{i=1}^n{F1_i}
\end{equation}

\section{Experiments and Results}
\label{Experiments and Results}
\label{sec:experiment}

\subsection{Baseline Models}
\label{baseline}
There are numerous effective methods have emerged globally for addressing VQA task. Among them, to match the ViOCRVQA dataset, we choose LoRRA, LaTr, PreSTU, and BLIP-2 according to the historical development of VQA methods and use them as baselines to perform experiments. Note that, these baselines were originally designed to support English, but we have adapted them to Vietnamese while still retaining their essence. Origin Latr and PreSTU were pre-trained on massive document images, scene text images, etc., and they demonstrated excellent performance in fine-tuning downstream tasks. However, Vietnamese is a low-resource language, so we can only fine-tune LaTr and PreSTU on the VQA task without doing pre-training, which requires a lot of computing resources and data.

\textbf{LoRRA}: Short for ``Look, Read, Reason and Answer'' \cite{singh2019towards}, this method uses a deep learning network to learn to combine information from various sources, including image and text features. This model has three main parts: VQA component helps reason about the answer based on the image. The Reading component helps the model read the text appearing in the image. The Answer module helps predict from an answer space or pointer to text read by the Reading component.

\textbf{LaTr}: ``Layout-Aware Transformer'' \cite{biten2022latr} based on the encoder-decoder transformer architecture (T5) \cite{raffel2020exploring} to build and includes three main blocks. The initial module focuses on a Language Model trained for document layouts with only text and layout details. The second module uses spatial embedding to embed scene text tokens and their positional information (OCR tokens bounding box). After pre-training, for the downstream task, the ViT \cite{dosovitskiy2020image} model is utilized to extract visual features. Finally, all these embedding features generated from these three modalities are used as input for the pre-trained transformer. The encoder then learns a representation that aligns the information from these modalities. This learned representation is later utilized by a decoder to analyze and generate an output, typically an answer.

\textbf{PreSTU}: ``Pre-Training for Scene-Text Understanding'' \cite{kil2023prestu} stands out from other models in tackling a common task by incorporating a unique approach. To ensure that models grasp spatial information in scene text and standardize the target output sequence during training, \cite{kil2023prestu} arranged OCR texts by their positions, sequentially from the top left to the bottom right. They then concatenated the sorted texts using a T5 separator </s>. Afterward, they randomly split the OCR texts into two segments: the first part is used as additional input, and the second part is used as the target. They then conduct pre-training on a huge amount of scene text image data. Finally, a pre-trained model that can ``understand scene text'' is fine-tuned for downstream tasks such as VQA, image captioning, etc.

\textbf{BLIP-2}: \citet{li2023blip} proposed a new novel approach to training vision-language model named ``Bootstrapping Language-Image Pre-training with Frozen Image Encoders and Large Language Models''. This approach uses CLIP \cite{radford2021learning} and freezes it like an image encoder which takes an image as input and extracts visual features. Then Querying Transformer (Q-Former), which is a lightweight transformer model, was trained to close the gap between the image encoder and the Large Language Model. This approach allows BLIP-2 to achieve competitive performance on vision-language tasks while requiring significantly less training compared to training everything from scratch.
\subsection{Experimental Configuration}

All baseline models and our proposed methods were trained and fine-tuned using the Adam optimization \cite{KingBa15}. We utilized an A100-GPU setup with 80GB of memory to train models, taking 10 hours on average for each method. We set the learning rate to 3e-05, dropout is set at 0.2, batch size is 32, and the training process is terminated after 5 epochs of not finding any growth in EM.

\subsection{Main Results}
Due to the nature of book genres often not being readily available on book covers and must synthesize information from the remaining fields. Therefore, we chose to omit them from the primary scope of OCR-VQA models discussed in this section. This decision enhances the efficiency of these models in addressing their core task.

\begin{table*}[htp]
\centering
\setlength{\tabcolsep}{10pt}
\caption{Results of our proposed methods and baselines test set.}
\label{table:kqua}
\begin{adjustbox}{max width=\textwidth}
\begin{tabular}{lrcc}
\hline
\textbf{Model}              & \textbf{Model Size} &   \textbf{EM (\%)}           &  \textbf{F1-score (\%)}           \\ \hline
LoRRA              &   26M              &   10.30                 &   21.54                 \\
BLIP-2              & 1.4B               &  21.45                &  55.23                 \\
LaTr               & 331M               &  30.80                 &  60.97                 \\
PreSTU             & 312M               &   33.86                &   66.25                 \\
\hline
\textbf{VisionReader\textsubscript{\textit{withBARTpho}} (ours)} & \textbf{220M}               &   \textbf{31.56}  &  \textbf{64.54}       \\
\textbf{VisionReader\textsubscript{\textit{withViT5}} (ours)}   & \textbf{315M}               &   {\textbf{41.16}} &  {\textbf{69.90}}\\
\hline
\end{tabular}
\end{adjustbox}
\end{table*}

Table \ref{table:kqua} shows that LoRRA, a pioneering VQA text-based approach, attained an EM of 10.30 and an F1-score of 21.54. While LoRRA can handle this dataset as a multi-label classification task, it still has many limitations that cause poor performance when the answer becomes free-from format in questions about book titles.

BLIP-2, a model boasting an impressive 1.4 billion parameters, showcased its prowess with an Exact Match (EM) score of 21.45 and an F1-score of 55.23. This performance, while commendable, was unexpectedly surpassed by LaTr, a model with a significantly smaller parameter count of just 331 million. LaTr achieved an EM of 30.80 and an F1-score of 60.97.

Even more intriguing was the performance of PeSTU, which, with a slightly smaller size than LaTr, managed to outperform all its predecessors with an EM of 33.86 and an F1-score of 66.25. This achievement serves as a poignant reminder that the path to optimizing machine learning models is multifaceted, involving more than just the accumulation of parameters.

One of our contributions is the VisionReader\textsubscript{\textit{withViT5}}, a model boasting 315 million parameters. Despite its relatively modest size, it has achieved the highest scores to date in both EM and F1-score metrics, with scores of 41.16 and 69.90, respectively, in the test set. This achievement underscores the effectiveness and potential of this our proposed method in tackling complex task like OCR-VQA.

Moreover, VisionReader\textsubscript{\textit{withBARTpho}}, equipped with only 220 million parameters, has established unprecedented benchmarks by surpassing all state-of-the-art baselines except PreSTU. VisionReader\textsubscript{\textit{withBARTpho}} demonstrates its remarkable performance when it achieved an EM of 31.56 and an F1-score of 64.54. This shows the efficacy of this proposed method in the OCR-VQA task while using fewer parameters.

\subsection{Results Book Genres}
\label{sub_gener}
In this section, we conducted individual experiments exclusively focusing on training the model using only the book genre field. The objective is to see behavior in synthesizing information from other fields on book covers in predicting genres.

\begin{table*}[h]
\centering
\caption{Results of book genres on ViOCRVQA dataset.}
\label{table:gerne}
\begin{tabular}{lllclc}
\hline
\textbf{Model}                    &  &  & \textbf{EM (\%)}     &  & \textbf{F1-score (\%)}     \\ \hline
LoRRA                    &  &  & 8.42 &  & 19.83 \\
BLIP-2                   &  &  & 37.89 &  & 47.83 \\
PreSTU                   &  &  & 51.17 &  & 61.96 \\
LaTr                     &  &  & 45.92 &  & 53.94 \\ \hline
VisionReader\textsubscript{\textit{withBARTpho}} (ours) &  &  & 47.78 &  & 56.53 \\
VisionReader\textsubscript{\textit{withViT5}} (ours)   &  &  & \textbf{57.24} &  & \textbf{64.41} \\ \hline
\end{tabular}
\end{table*}

From the genre prediction results in Table \ref{table:gerne}, the VisionReader\textsubscript{\textit{withViT5}} model achieved the best performance in the task of predicting book genre with an EM accuracy score of 57.24. PreSTU ranked second with an EM of 51.17, while VisionReader\textsubscript{\textit{withBARTpho}} ranked third with an EM of 47.78. These results underscore the effectiveness of our proposed method in synthesizing information across various fields to enhance genre prediction accuracy on book covers.

\section{Result Analysis}
\label{Result Analysis}
\subsection{Detail Results}
To give an in-depth analysis of how baselines and our proposed methods achieved their performance on the ViOCRVQA dataset, we categorized the answers given by these models into four aspects relevant to the information their respective questions inquiry: title, author, publisher, and translator. Detailed results are shown in Table \ref{detail_results}.
\begin{table*}[h]
\centering
\setlength{\tabcolsep}{10pt}
\centering
\caption{Detail results of baseline models and proposed method on different fields.}
\label{detail_results}
\begin{adjustbox}{max width=\textwidth}
\begin{tabular}{lcccccccc}
\hline
       & \multicolumn{2}{c|}{Title}            & \multicolumn{2}{c|}{Author}           & \multicolumn{2}{c|}{Publisher}        & \multicolumn{2}{c}{Translator} \\ \hline 
Model & EM (\%) & \multicolumn{1}{c|}{F1-score (\%)} & EM (\%) & \multicolumn{1}{c|}{F1-score (\%)} & EM (\%) & \multicolumn{1}{c|}{F1-score (\%)} & EM (\%) & F1-score (\%) \\ \hline
LoRRA    & 3.58 & \multicolumn{1}{c|}{20.67} & 11.88 & \multicolumn{1}{c|}{21.47} & 17.85 & \multicolumn{1}{c|}{21.74} & 9.42         & 23.08        \\ 
BLIP-2   & 3.49 & \multicolumn{1}{c|}{51.40} & 22.89 & \multicolumn{1}{c|}{52.34} & 48.76 & \multicolumn{1}{c|}{77.34} & 20.56 & 49.34        \\
PreSTU & 9.58 & \multicolumn{1}{c|}{63.09} & 25.39 & \multicolumn{1}{c|}{48.14} & 53.30 & \multicolumn{1}{c|}{79.23} & 21.53         & 44.46        \\
LaTr   & 4.90 & \multicolumn{1}{c|}{52.18} & 28.19 & \multicolumn{1}{c|}{51.20} & 61.60 & \multicolumn{1}{c|}{84.03} & 24.98 & 49.45        \\
\hline
VisionReader\textsubscript{\textit{withBARTpho}} (ours)    & 7.55 & \multicolumn{1}{c|}{59.59} & 31.50 & \multicolumn{1}{c|}{57.20} & 57.31 & \multicolumn{1}{c|}{81.90} & 27.36         & 51.61        \\ 
VisionReader\textsubscript{\textit{withViT5}} (ours)    & \textbf{13.42} & \multicolumn{1}{c|}
{\textbf{64.34}} & \textbf{44.19} & \multicolumn{1}{c|}{\textbf{64.29}} & \textbf{65.73} & \multicolumn{1}{c|} {\textbf{85.71}} & \textbf{41.58} & \textbf{58.09} \\
\hline
\end{tabular}
\end{adjustbox}
\end{table*}

\begin{figure*}[htp]
\centering
\includegraphics[width=0.9\textwidth,height=4.5cm]{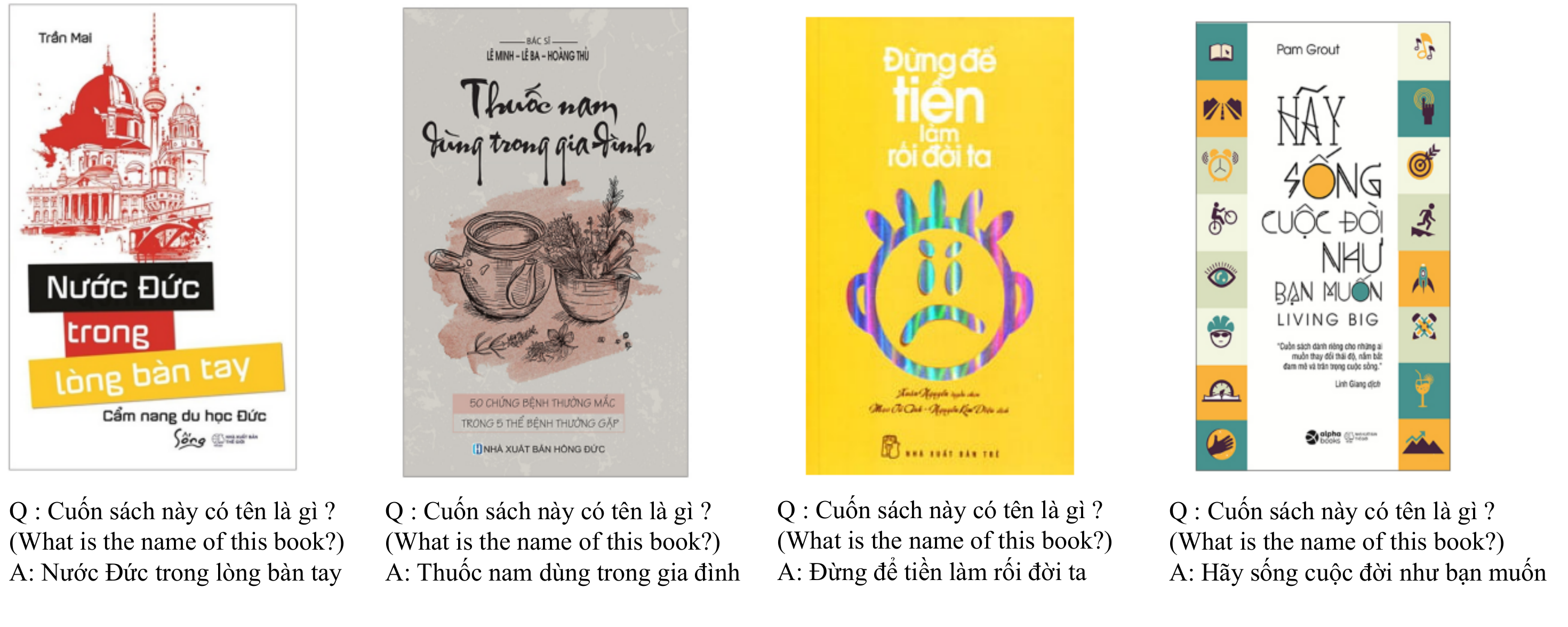}
\caption{Several examples of question-answer pairs related to title, title with unusual fonts, and messy arrangements.}
\label{fig:titlee}
\end{figure*}

\textbf{Title:} Table \ref{detail_results} demonstrates that providing answers to questions in the book title becomes more complex than in other fields, especially as the model shows the lowest performance in EM. The main reason for this phenomenon is that titles often include complex text structures and use many different typefaces. This diversity makes it more difficult for models to accurately extract and understand information from titles (see Figure \ref{fig:titlee}). This poses a significant challenge in providing answers to questions related to book titles.

\begin{figure*}[htp]
\centering
\includegraphics[width=0.9\textwidth,height=4.5cm]{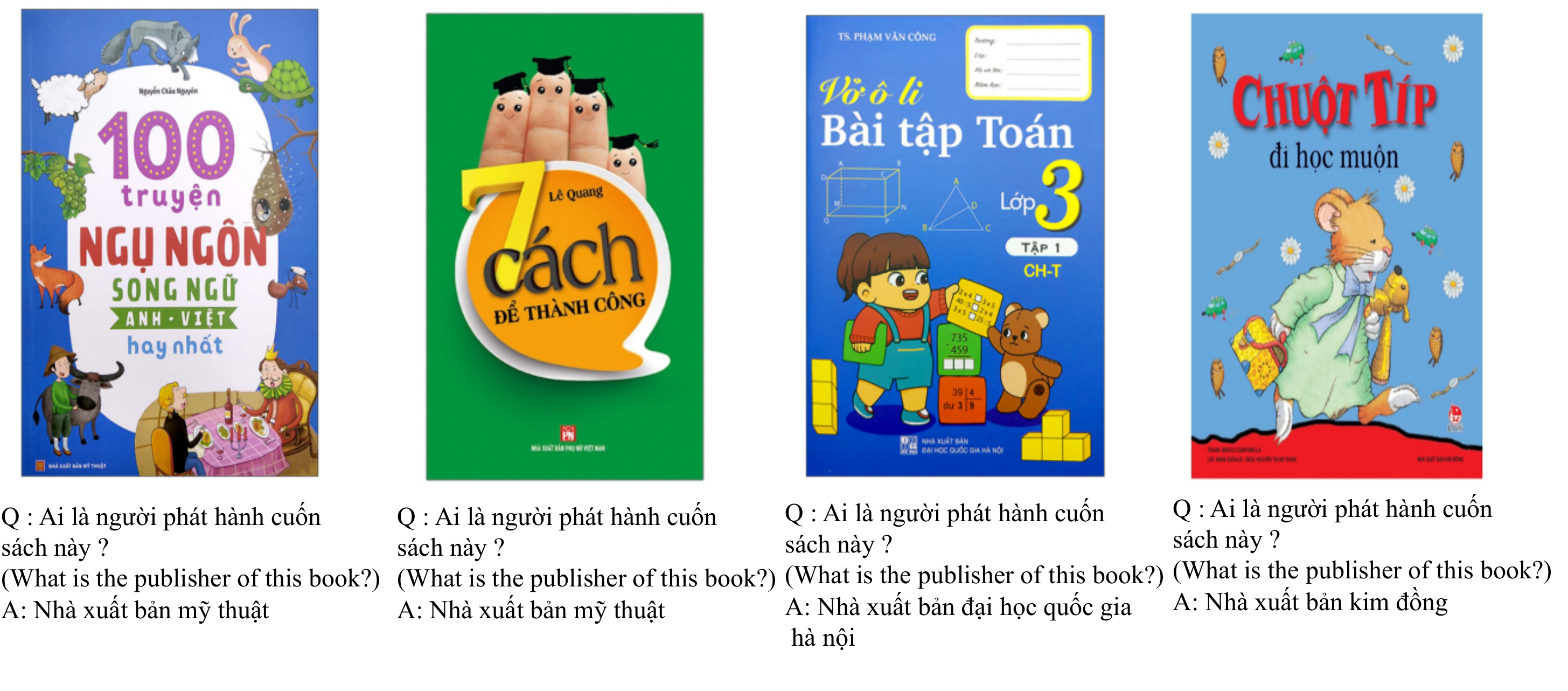}
\caption{Several examples of question-answer pairs are related to the publisher, the publisher's name is often placed at the bottom of the book cover.}
\label{fig:publisher}
\end{figure*}

\textbf{Publisher:} In Table \ref{detail_results}, the results show that the field related to publishers yielded the highest EM and F1-score. Given the relatively limited number of publishers, which stands at 176, answering questions about publishers appears to be more straightforward than other fields. Moreover, publishers' names are often prominently displayed and clearly identified on the back cover of books, facilitating easier recognition (see Figure \ref{fig:publisher}). This prominence underscores the potential of publisher-related information in enhancing model performance.

\begin{figure*}[htp]
\centering
\includegraphics[width=0.9\textwidth,height=4.5cm]{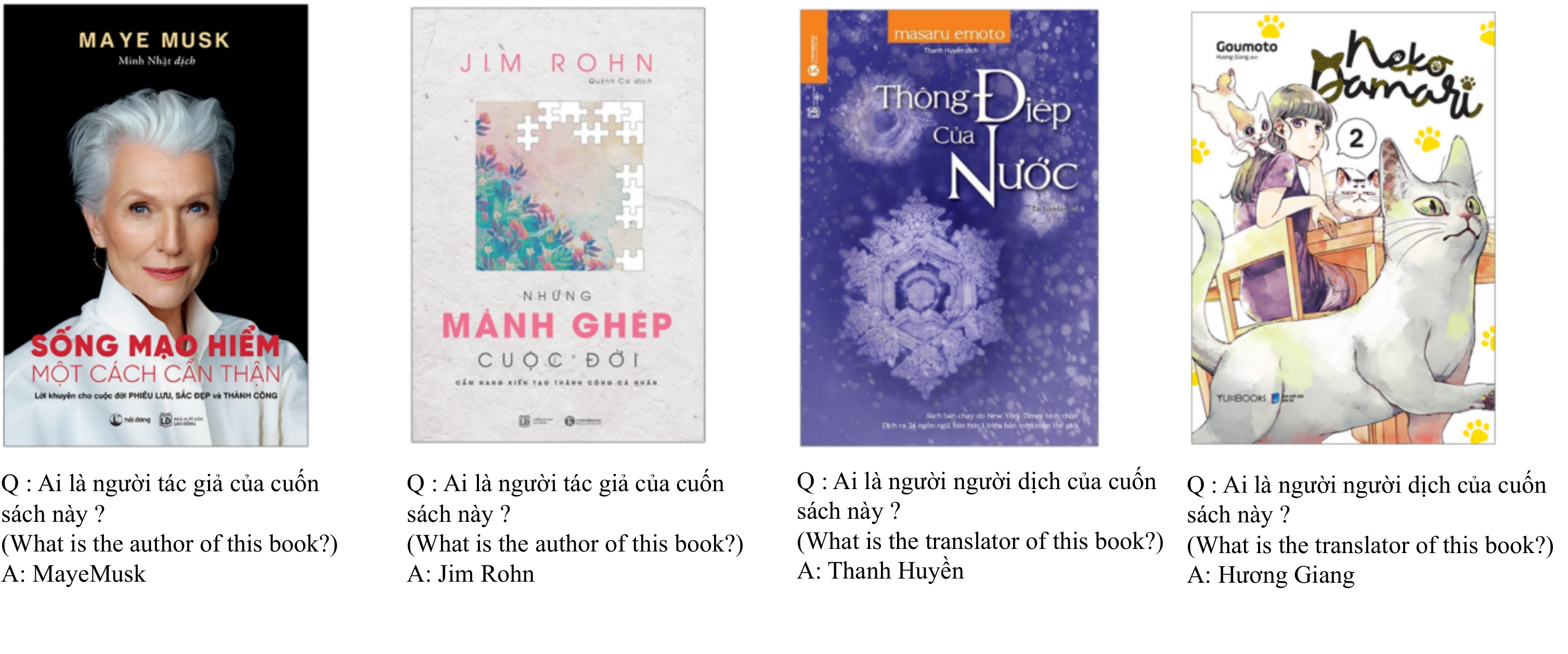}
\caption{Several examples of question-answer pairs are related to the author and translator, the author and translator is often placed near each other.}
\label{fig:authorandtranslator}
\end{figure*}

\textbf{Author and Translator:} On book covers, the author and translator are usually placed next to each other, with the author's name often printed larger than the translator's name (see Figure \ref{fig:authorandtranslator}). This small difference in representation between the author's name and the translator's name facilitates the model in identifying and providing more accurate answers. This is clearly illustrated in Table \ref{detail_results}, where the results for the author field are generally higher than for the translator field.

\clearpage

\subsection{Impact of OCR System Performance}
We evaluated the performance of the OCR system SwinTextSpotter \cite{huang2022swintextspotter} to assess its impact on the efficiency of various models. To accomplish this, we segmented the test data based on the proportion of text successfully identified by the OCR system relative to the total amount of text in the answer. These segments included text with successful recognition rates of 25\%, 50\%, 75\%, and 100\% of the total text in answer. 
\begin{figure}[htp]
  \centering
\includegraphics[width=0.49\textwidth]{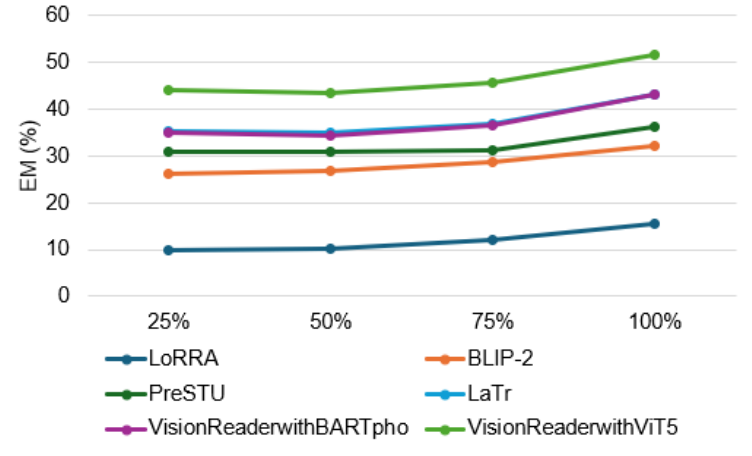}
\includegraphics[width=0.49\textwidth]{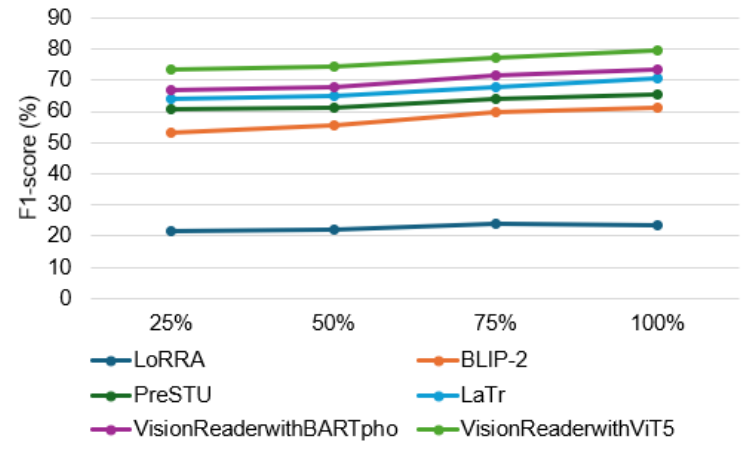}
  \caption{EM and F1-score with the percentage of text in the answer detected by the OCR system}
  \label{fig:OCR}
\end{figure}

The performance of both baseline and our models in Figure \ref{fig:OCR} follows a consistent trend: initially dipping slightly as the answer token ratio in OCR text ranges from 25\% to 50\%, then gradually increasing. Notably, there is  a significant boost in performance when the token ratio hits 100\%. Despite differences in models, all show similar F1-score patterns, indicating that as more text is detected by the OCR system, performance improves. This underscores the crucial role of OCR text in providing context and supporting data for accurate predictions by the VQA model.

This highlights a significant observation that even when all answer tokens are accurately identified by OCR system, the performance metrics for EM and F1-score only reach moderately acceptable levels, remaining below 55.00\% and 80.00\%, respectively. This underscores the notable challenge encountered by VQA models when dealing with the ViOCRVQA dataset.

\subsection{Is Object Necessary for OCR Visual Question Answering?}
\label{Ablation Study}

As mentioned before in Section \ref{metho}, we assumed that information about the object plays an important role in defining the question to determine which answer is reasonable. To prove this, we conducted specific experiments on our proposed methods by removing object features.

\begin{table*}[htp]
\centering
\setlength{\tabcolsep}{10pt}
\centering\caption{Performance when removing the object feature in our proposed methods. $\triangle$ denotes the increase (↑) and decrease (↓) in the performance of our method compared to using object features.} 
\label{no_obj}
\begin{adjustbox}{width=1\textwidth}
\begin{tabular}{ccc|cc|cc|cc}
\hline
 & \multicolumn{2}{c|}{Title} & \multicolumn{2}{c|}{Author} & \multicolumn{2}{c|}{Publisher} & \multicolumn{2}{c}{Translator} \\ \hline
 Model & EM (\%)           & F1-score (\%)          & EM (\%)           & F1-score (\%)           & EM  (\%)            & F1-score (\%)           & EM (\%)             & F1-score (\%)            \\ \hline
\begin{tabular}[c]{@{}c@{}}VisionReader\textsubscript{\textit{withBARTpho}}\\ $\triangle$\end{tabular} &
  \begin{tabular}[c]{@{}c@{}}4.95\\ \textcolor{red}{$\downarrow 2.60$}\end{tabular} &
  \begin{tabular}[c]{@{}c@{}}47.72\\ \textcolor{red}{$\downarrow 11.87$}\end{tabular} &
  \begin{tabular}[c]{@{}c@{}}25.08\\ \textcolor{red}{$\downarrow 6.42$}\end{tabular} &
  \begin{tabular}[c]{@{}c@{}}47.99\\ \textcolor{red}{$\downarrow 9.21$}\end{tabular} &
  \begin{tabular}[c]{@{}c@{}}58.56\\ \textcolor{blue}{$\uparrow 1.25$}\end{tabular} &
  \begin{tabular}[c]{@{}c@{}}83.08\\ \textcolor{blue}{$\uparrow 1.18$}\end{tabular} &
  \begin{tabular}[c]{@{}c@{}}20.4\\ \textcolor{red}{$\downarrow 6.96$}\end{tabular} &
  \begin{tabular}[c]{@{}c@{}}42.14\\ \textcolor{red}{$\downarrow 9.47$}\end{tabular} \\
\begin{tabular}[c]{@{}c@{}}VisionReader\textsubscript{\textit{withViT5}}\\ $\triangle$\end{tabular} &
  \begin{tabular}[c]{@{}c@{}}6.81\\ \textcolor{red}{$\downarrow 6.61$}\end{tabular} &
  \begin{tabular}[c]{@{}c@{}}48.63\\ \textcolor{red}{$\downarrow 15.71$}\end{tabular} &
  \begin{tabular}[c]{@{}c@{}}31.43\\ \textcolor{red}{$\downarrow 12.76$}\end{tabular} &
  \begin{tabular}[c]{@{}c@{}}53.14\\ \textcolor{red}{$\downarrow 11.15$}\end{tabular} &
  \begin{tabular}[c]{@{}c@{}}68.28\\ \textcolor{blue}{$\uparrow 2.55$}\end{tabular} &
  \begin{tabular}[c]{@{}c@{}}87.17\\ \textcolor{blue}{$\uparrow 1.46$}\end{tabular} &
  \begin{tabular}[c]{@{}c@{}}26.17\\ \textcolor{red}{$\downarrow 15.41$}\end{tabular} &
  \begin{tabular}[c]{@{}c@{}}46.04\\ \textcolor{red}{$\downarrow 12.05$}\end{tabular}
\\
\hline
\end{tabular}
\end{adjustbox}
\label{no_obj}
\end{table*}

Table \ref{no_obj} illustrates a discernible downward trend in the efficacy of our proposed methods across various domains is observable upon the removal of object features, with the exception of the publisher field. Moreover, it is worth noting that while there was a slight increase in the score for the publisher field, it was not significant. Thus, if the aim is to boost the performance of the model in general, it becomes evident that the object plays a crucial role. Its inclusion not only enhances question understanding but also enriches the context necessary for the model to generate accurate answers.

\subsection{How Does OCR Contribute to Understanding Vietnamese Text in Images?}
By conducting experiments on removing OCR, we sought to shed light on the significance of OCR and its direct impact on the overall performance of OCR-VQA models in our ViOCRVQA dataset. Moreover, this analysis serves to emphasize the importance of OCR as a means to enhance the efficacy of OCR-VQA task. 

\begin{table*}[htp]
\centering
\setlength{\tabcolsep}{10pt}
\caption{Performance when removing the OCR in our proposed method. $\triangle$ denotes the increase (↑) and decrease (↓) in the performance of our method compared to the use of the OCR feature and OCR text.}
\begin{adjustbox}{width=\textwidth}
\begin{tabular}{ccc|cc|cc|cc}
\hline
 & \multicolumn{2}{c|}{Title} & \multicolumn{2}{c|}{Author} & \multicolumn{2}{c|}{Publisher} & \multicolumn{2}{c}{Translator} \\ \hline
 Model & EM (\%)           & F1-score (\%)          & F1-score (\%)           & F1-score (\%)           & EM (\%)             & F1-score (\%)            & EM (\%)             & F1-score (\%)            \\ \hline
\begin{tabular}[c]{@{}c@{}}VisionReader\textsubscript{\textit{withBARTpho}}\\ $\triangle$\end{tabular} &
  \begin{tabular}[c]{@{}c@{}}0.5\\ \textcolor{red}{$\downarrow 7.05$}\end{tabular} &
  \begin{tabular}[c]{@{}c@{}}17.72\\ \textcolor{red}{$\downarrow 41.87$}\end{tabular} &
  \begin{tabular}[c]{@{}c@{}}13.09\\ \textcolor{red}{$\downarrow 18.41$}\end{tabular} &
  \begin{tabular}[c]{@{}c@{}}17.28\\ \textcolor{red}{$\downarrow 29.92$}\end{tabular} &
  \begin{tabular}[c]{@{}c@{}}28.68\\ \textcolor{red}{$\downarrow 28.63$}\end{tabular} &
  \begin{tabular}[c]{@{}c@{}}62.15\\ \textcolor{red}{$\downarrow 19.75$}\end{tabular} &
  \begin{tabular}[c]{@{}c@{}}12.23\\ \textcolor{red}{$\downarrow 15.13$}\end{tabular} &
  \begin{tabular}[c]{@{}c@{}}10.14\\ \textcolor{red}{$\downarrow 36.37$}\end{tabular} \\
\begin{tabular}[c]{@{}c@{}}VisionReader\textsubscript{\textit{withViT5}}\\ $\triangle$\end{tabular} &
  \begin{tabular}[c]{@{}c@{}}0.70\\ \textcolor{red}{$\downarrow 12.72$}\end{tabular} &
  \begin{tabular}[c]{@{}c@{}}19.29\\ \textcolor{red}{$\downarrow 45.05$}\end{tabular} &
  \begin{tabular}[c]{@{}c@{}}16.54\\ \textcolor{red}{$\downarrow 27.65$}\end{tabular} &
  \begin{tabular}[c]{@{}c@{}}20.22\\ \textcolor{red}{$\downarrow 44.07$}\end{tabular} &
  \begin{tabular}[c]{@{}c@{}}38.20\\ \textcolor{red}{$\downarrow 27.53$}\end{tabular} &
  \begin{tabular}[c]{@{}c@{}}73.20\\ \textcolor{red}{$\downarrow 12.51$}\end{tabular} &
  \begin{tabular}[c]{@{}c@{}}10.23\\ \textcolor{red}{$\downarrow 31.35$}\end{tabular} &
  \begin{tabular}[c]{@{}c@{}}15.24\\ \textcolor{red}{$\downarrow 42.85$}\end{tabular}
\\
\hline
\end{tabular}
\end{adjustbox}
\label{no_ocr}
\end{table*}

As depicted in Table \ref{no_ocr}, the lack of OCR results in significantly reduced performance for both VisionReader\textsubscript{\textit{withViT5}} and VisionReader\textsubscript{\textit{withBARTpho}}. When not using OCR, the score drops very seriously, especially in the title field, the EM drops close to the lower border and the F1-score drops by two-thirds. This shows the importance of OCR in the OCR-VQA task. It can be said that improving and ensuring OCR performance is one of the best ways to increase and ensure performance in OCR-VQA task.

\subsection{How Do Answer and Question Lengths Affect Model Performance?}
\begin{table}[htp]
  \centering
  \caption{Group of answer and question length in test set.}
  \begin{tabular}{lrrr}
    \hline
    \textbf{Group} & \textbf{Length (n)} & \textbf{Answer Samples} &\textbf{Question Samples} \\
    \hline
    Short & $n \leq 5$ & 12637 & 12531\\
    Medium & $5 < n \leq 10$ & 4392 & 4894\\
    Long & $10 < n \leq 15$ &1215 & 1131\\
    Very long & $n > 15$& 357 &43\\
    \hline
  \end{tabular}
  \label{ans_ques_len}
\end{table}

We divided questions and answers based on their length in the number of tokens followed in the study of \citet{nguyen2023openvivqa}. The details of the test set based on different groups of lengths are in Table \ref{ans_ques_len}. Classification is done as follows:

\begin{itemize}
\item Short question (and short answer): These are questions and answers shorter than 6 tokens.

\item Medium question (and medium answer): This group includes questions and answers from 6 to 10 tokens.

\item Long question (and long answer): Questions and answers in this group from 11 to 15 tokens. 

\item Very long question (and very long answer): This group contains questions and answers from 16 tokens and longer.
\end{itemize}

\begin{figure}[htp]
  \centering
\includegraphics[width=0.49\textwidth]{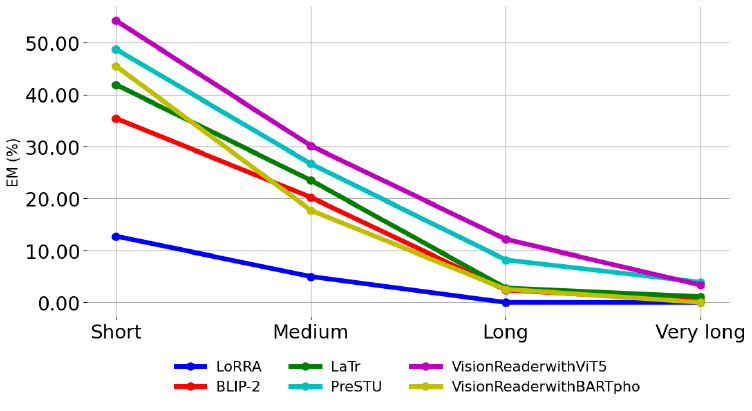}
\includegraphics[width=0.49\textwidth]{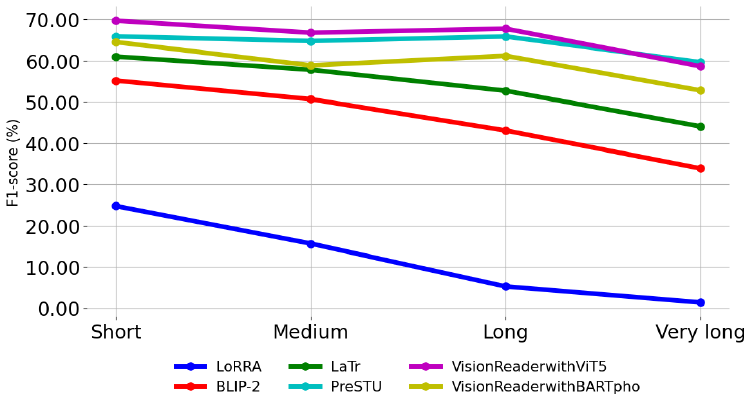}
  \caption{The results of models based on answer length.}
  \label{answer length}
\end{figure}

The illustration from Figure \ref{answer length} demonstrates the performance of the model for different answer lengths. The results indicate that the length of the answer plays a crucial role in affecting the performance of models. For the EM metric, short answers are ideal for the model, as the EM decreases rapidly with increasing answer length. In addition to that, for the F1-score, the performance of the models remains consistently good across a range of answer lengths but struggles to achieve good results with very long answers.

\begin{figure}[htp]
  \centering
\includegraphics[width=0.49\textwidth]{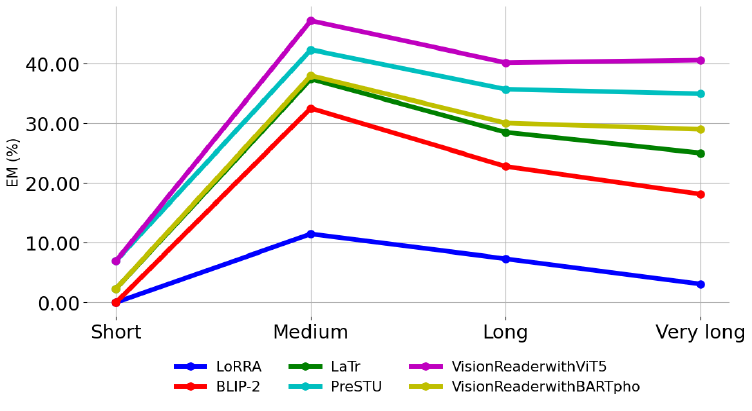}
\includegraphics[width=0.49\textwidth]{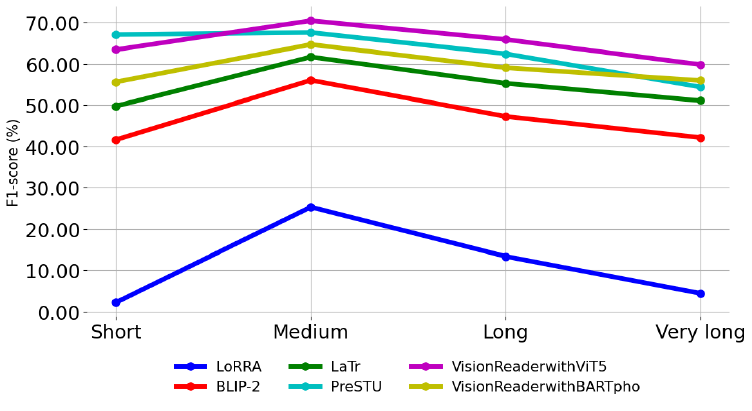}
  \caption{The results of models based on question length.}
  \label{question length}
\end{figure}

The performance illustration of the model with question length in Figure \ref{question length} indicates that both overly short and long questions do not yield high F1-score and EM. With overly short questions, there is not sufficient information provided for the model to produce high-quality results. Similarly, overly very long questions result in diluted input information, making it unclear what information needs to be extracted, hence leading to lower scores. Questions of medium to long length, however, yield positive results.

\subsection{How large a dataset is enough?}
To better understand how data size affects model performance, we conducted a series of experiments in which the model was trained with different percentages of the dataset: 25\%, 50\%, 75\%, and 100\%.

Through these experiments, we observe a clear positive trend: model performance increases steadily with the size of the dataset. Specifically, when increasing from 25\% to 50\%, then 75\%, and finally to 100\% of the training dataset, we observe a continuous improvement in performance.
\begin{figure*}[htp]
  \centering
  \includegraphics[width=0.8\textwidth]{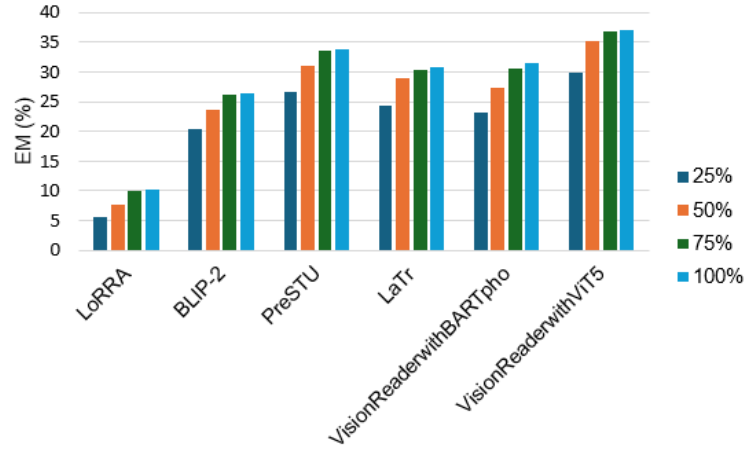}
  \centering\caption{The EM results of models based on the percentage of training data}
  \label{fig:Em_all}
\end{figure*}
\begin{figure*}[htp]
  \centering
  \includegraphics[width=0.8\textwidth]{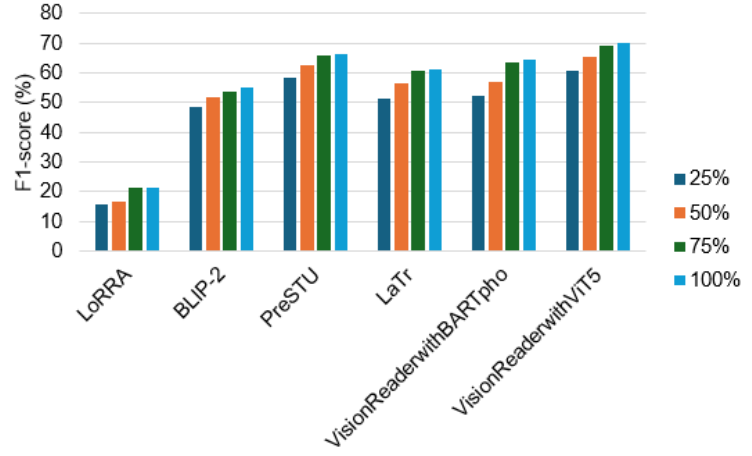}
  \centering\caption{The F1-score results of models based on the percentage of training data}
  \label{fig:F1_all}
\end{figure*}


As the number of training data increases, Figure \ref{fig:Em_all} and Figure \ref{fig:F1_all} illustrate a corresponding improvement in the results achieved. The results showed a significant increase from 25\% to 50\%, followed by a more modest increase from 50\% to 75\%. In particular, the increase becomes very small from 75\% to 100\%. As the data stream continues to increase, VQA models gradually reach a saturation point. The larger ViOCRVQA dataset can indeed enhance the results of models but not significantly, instead, let to improve the model to enhance performance.

Data is often considered the deciding factor in improving the performance of machine learning models. In general, increasing the size of training data can yield a significant improvement in performance, but beyond a certain threshold, this improvement may no longer be significant. In this context, the ViOCRVQA dataset has demonstrated that it provides a sufficiently rich amount of data to effectively train models.

\section{Conclusion and Future Work}
\label{Conclusion and Future Work}
In this article, we presented the ViOCRVQA dataset, which includes 28,282 images and 123,781 question-answer pairs. This dataset will be publicly shared with the research community, especially with the Vietnamese research community, and will become the largest dataset serving the task of VQA in Vietnamese. In addition, we also analyzed, explored, and conducted experiments on state-of-the-art (SOTA) models and discovered their limitations when applied to the ViOCRVQA dataset. Furthermore, we developed the VisionReader, which is optimized to solve OCR-VQA task in Vietnamese. Our proposed methods proved to be superior to current SOTA models, opening up new and more effective approaches for the research community in this field.

Through our experiments, we discovered the undeniable importance of optical character recognition (OCR) systems in the OCR-VQA task. Although this task mainly focuses on extracting textual information from images, we have demonstrated that the relationship between objects in images and textual information is intimate. Our proposed model achieved significant performance improvement by using information about objects in images. This discovery opens up a new approach to enhance the quality and accuracy of other OCR-VQA models in the future.

For future work, we will leverage large vision models such as visualGPT \cite{chen2022visualgpt} visionLLM \cite{wang2024visionllm} and large language models such as GPT-3 \cite{brown2020language}, LLaMA \cite{touvron2023llama}, etc. to enhance the performance of the OCR-VQA model. Thanks to their ability to represent features well, these models will help us better understand the relationship between text and images. Besides, we also propose to conduct experiments with different OCR systems such as FOTS \cite{liu2018fots}, Mask Textspoter \cite{lyu2018mask}, etc. to compare and evaluate their performance. Another potential approach is to combine OCR with VQA models to make them multitasking models, where one model can simultaneously recognize text and answer questions about images. Finally, we would also want to conduct further research on the possibility of applying reinforcement learning techniques to improve the quality of the OCR-VQA model on the ViOCRVQA dataset.

\section*{Acknowledgements}
This research is funded by Vietnam National University HoChiMinh City (VNU-HCM) under the grant number DS2024-26-01.

\section*{Author Contributions Statement}
 \textbf{Huy Quang Pham:} Conceptualization; Formal analysis; Investigation; Methodology; Validation; Visualization; Writing - review \& editing. \textbf{Thang Kien-Bao Nguyen:} Conceptualization; Data curation; Formal analysis; Investigation; Validation; Visualization; Writing - original draft. \textbf{Quan Van Nguyen:} Conceptualization; Data curation; Investigation; Methodology; Writing - original draft. \textbf{Dan Quang Tran:} Conceptualization; Data curation; Investigation; Methodology; Writing - original draft. \textbf{Nghia Hieu Nguyen:} Conceptualization; Data curation; Investigation; Methodology; Writing - original draft. \textbf{Kiet Van Nguyen:} Conceptualization; Formal analysis; Investigation; Methodology; Validation; Supervision; Writing - review \& editing. \textbf{Ngan Luu-Thuy Nguyen:} Conceptualization; Formal analysis; Investigation; Methodology; Validation; Supervision; Writing - review \& editing.

 \section*{Declarations}

\textbf{Conflict of interest} The authors declare that they have no conflict of interest.

\section*{Data Availability}

Data will be made available on reasonable request.

\begin{appendices}
\section*{Appendix}
\subsection*{Several understanding about the dataset ViOCRVQA}


\begin{figure*}[htp]
\centering
\includegraphics[width=0.9\textwidth, height=6cm]{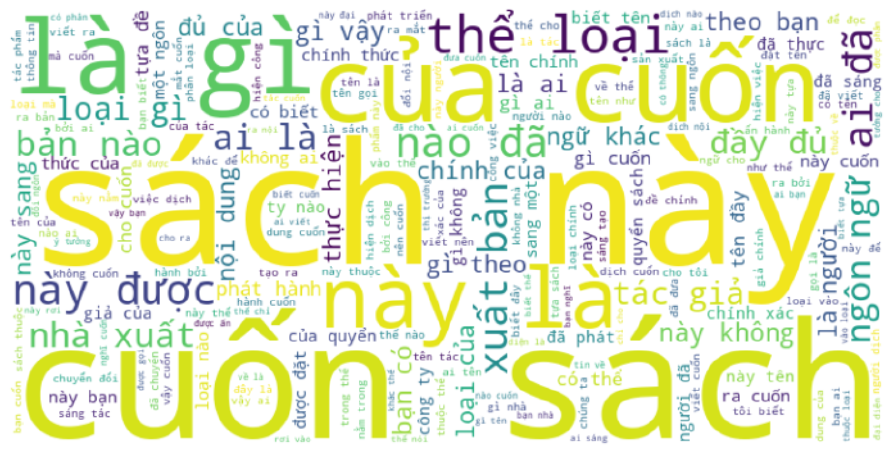}
\caption{Question word cloud.}
\end{figure*}
\begin{figure*}[h]
\centering
\includegraphics[width=0.9\textwidth, height=6cm]{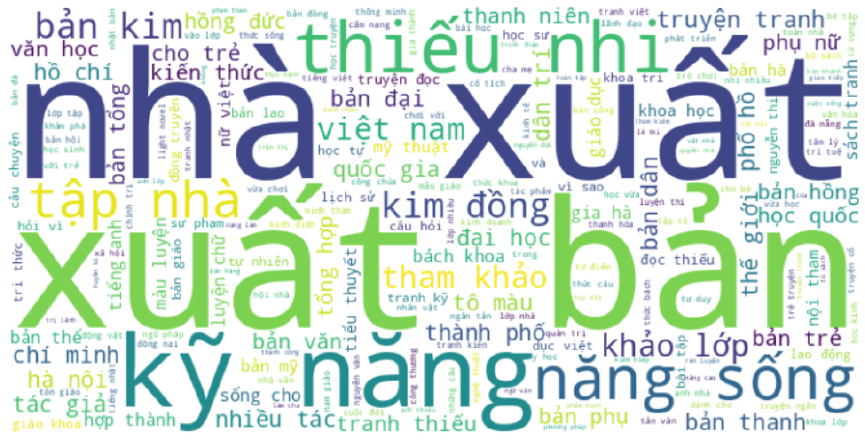}
\caption{Answer word cloud.}
\end{figure*}
\begin{table}[h]
\caption{The proportion of books based on genre.}
\begin{tabular}{lrr}
\hline
\multicolumn{1}{c}{Type}                                          & Count & Ratio (\%) \\ \hline
Lịch sử   - Địa lý - Tôn Giáo (History - Geography - Religion)    & 701   & 2.48            \\
Tâm lý - Kĩ năng Sống (Psychology - Life   Skills)                & 2,834 & 10.02           \\
Hề hước (Comic)                                                             & 2,089 & 7.39            \\
Giáo   khoa - Tham khảo (Textbook - Reference)                    & 4,323 & 15.29           \\
Văn   học (Literature)                                            & 3,603 & 12.74           \\
Kinh   Tế (Economics)                                             & 1,039 & 3.67            \\
Nữ   công gia chánh (Housewife)                                   & 246   & 0.87            \\
Khoa   học kỹ thuật (Science Technology)                          & 770   & 2.72            \\
Sách   học Ngoại Ngữ (Foreign Language Learning Books)            & 1,462 & 5.17            \\
Thiếu   Nhi (Children)                                            & 8,882 & 31.41           \\
Âm   nhạc - Mỹ thuật - Thời trang (Music - Art - Fashion)         & 68    & 0.24            \\
Giáo   trình (Curriculum)                                         & 23    & 0.08            \\
Nuôi   dạy con (Raise up child)                                   & 561   & 1.98            \\
Từ   điển (Dictionary)                                            & 155   & 0.55            \\
Tiểu   sử - Hồi ký (Biography - Memoirs)                          & 301   & 1.06            \\
Chính   trị - Pháp lý - Triết Học (Politics - Legal - Philosophy) & 377   & 1.33            \\
Khác   (Other)                                                    & 846   & 2.99            \\ \hline
\end{tabular}
\end{table}
\clearpage


\subsection*{Example Results of Models}
\begin{figure*}[htp]
  \centering
  \includegraphics[width=0.9\textwidth, height=7cm]{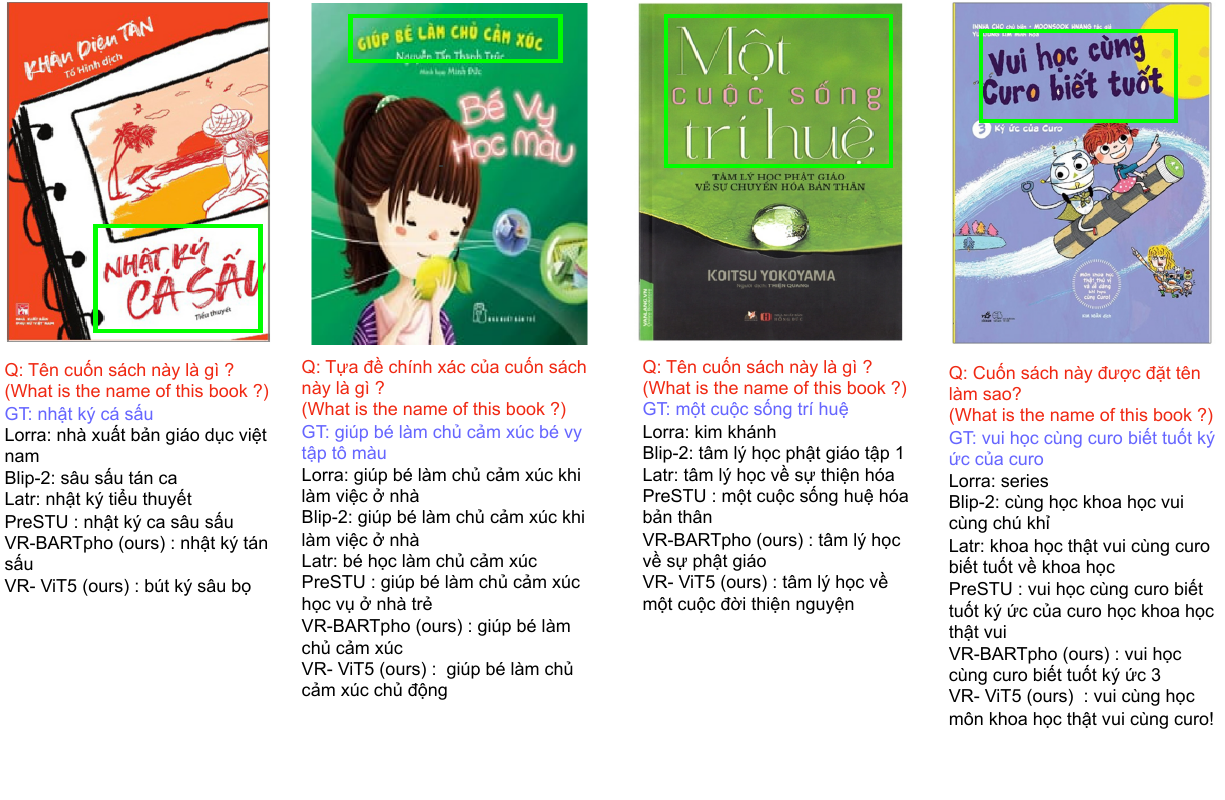}
  \caption{Examples results of models where book titles have unusual fonts. (Note: VR-BARTpho is VisionReader\textsubscript{\textit{withBARTpho}}, VR-ViT5 is VisionReader\textsubscript{\textit{withViT5}})}
  \end{figure*}
  \begin{figure*}[!h]
  \centering
  \includegraphics[width=0.9\textwidth, height=7cm]{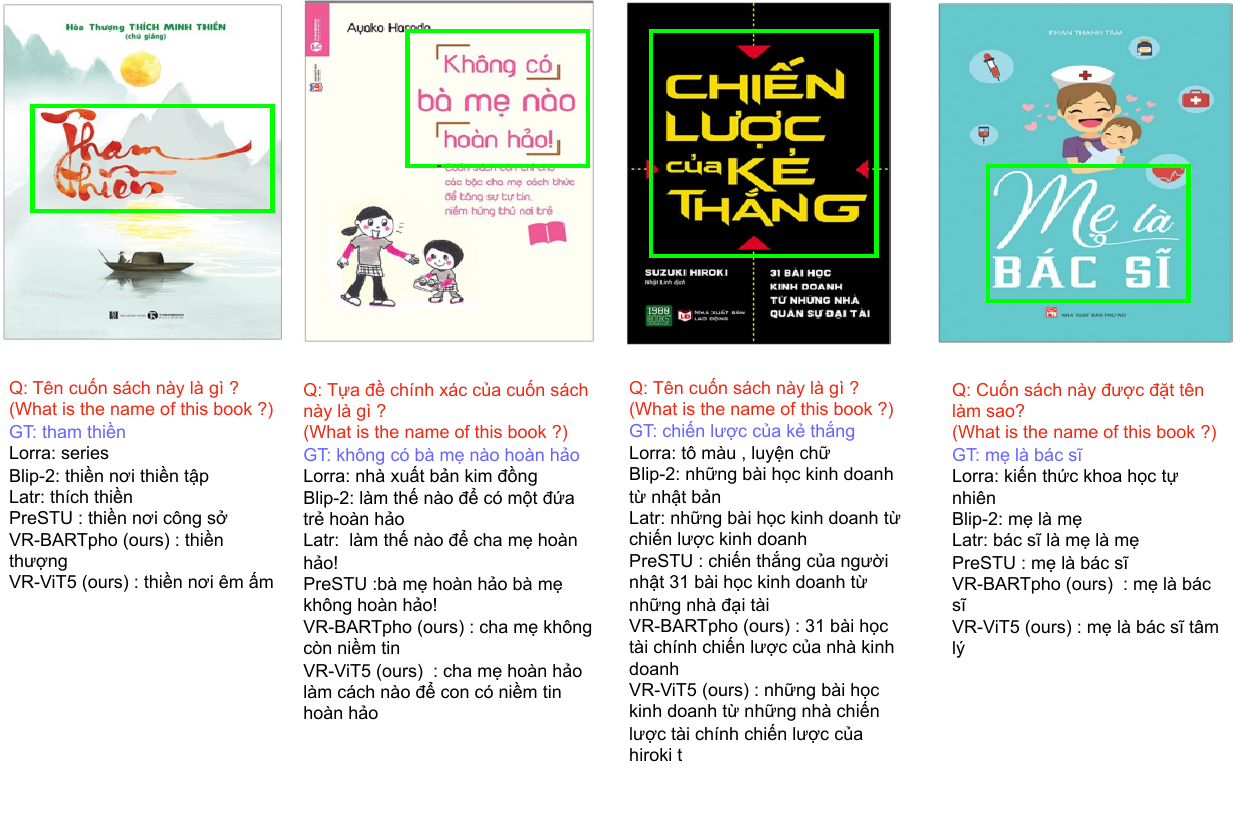}
  \caption{Examples results of models in title field. (Note: VR-BARTpho is VisionReader\textsubscript{\textit{withBARTpho}}, VR-ViT5 is VisionReader\textsubscript{\textit{withViT5}})}
  \end{figure*}
  \begin{figure*}[h]
  \centering
  \includegraphics[width=0.9\textwidth, height=7cm]{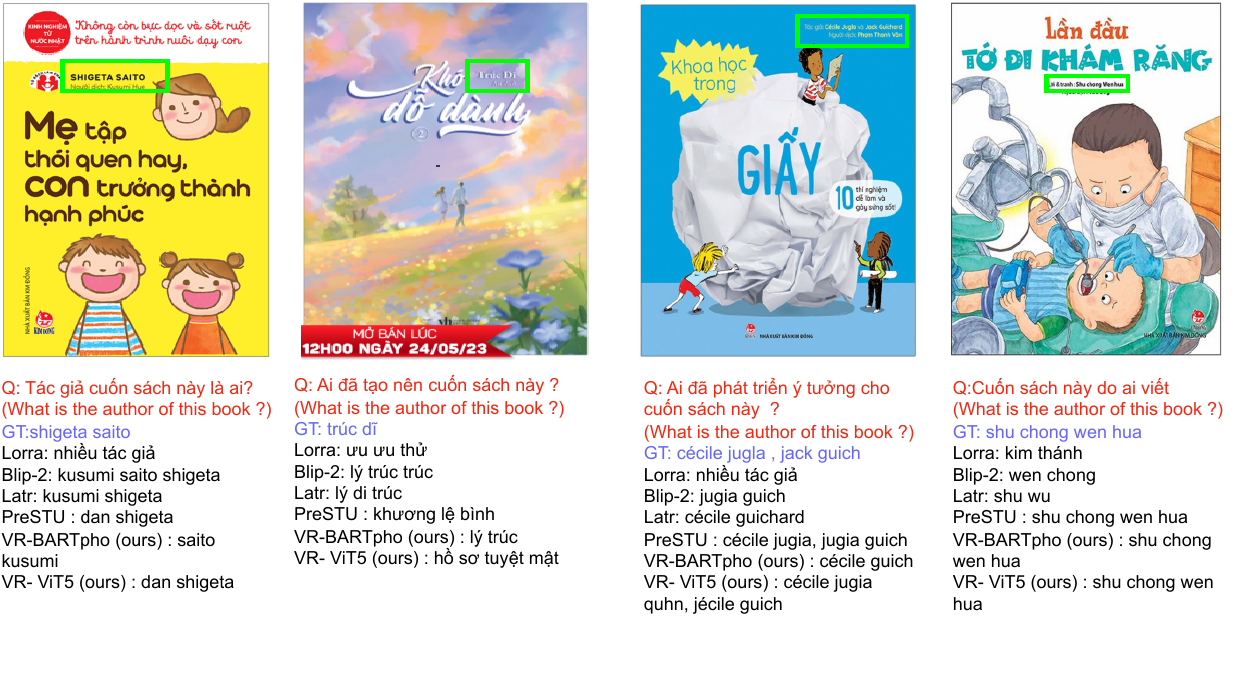}
  \caption{Examples results of models in author field (Note: VR-BARTpho is VisionReader\textsubscript{\textit{withBARTpho}}, VR-ViT5 is VisionReader\textsubscript{\textit{withViT5}})}
\end{figure*}

  \begin{figure*}[h]
  \centering
  \includegraphics[width=0.9\textwidth, height=7cm]{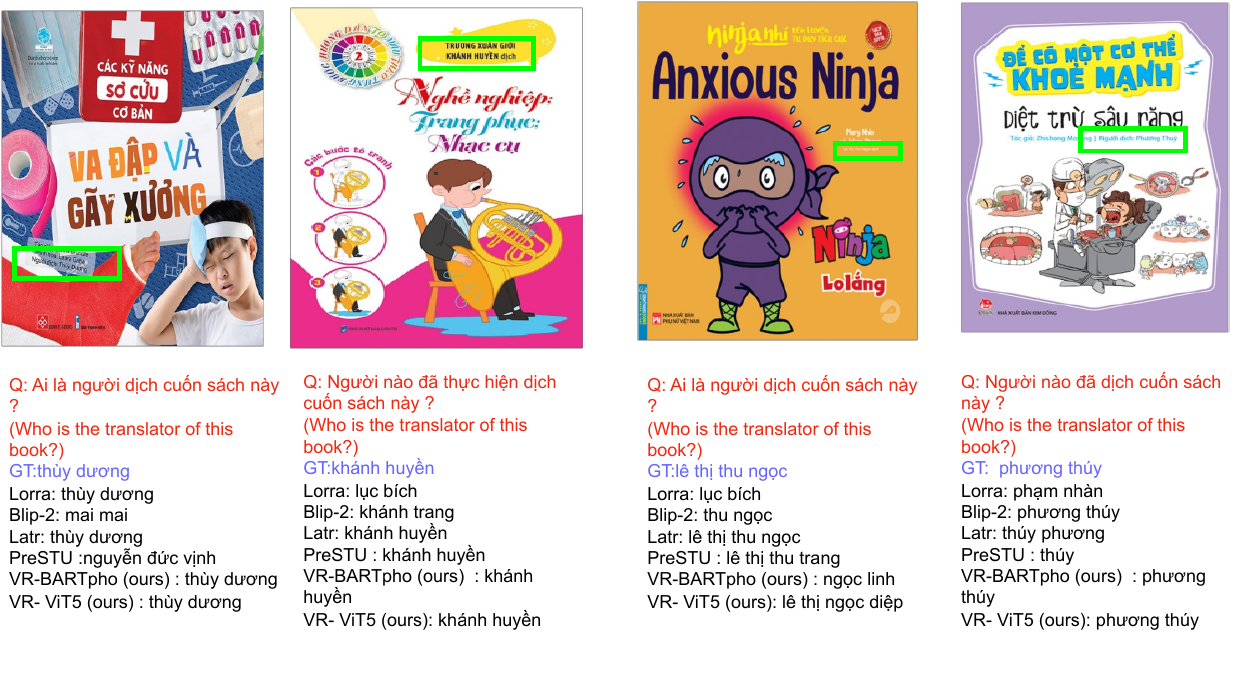}
  \caption{Examples results of models in translator field. (Note: VR-BARTpho is VisionReader\textsubscript{\textit{withBARTpho}}, VR-ViT5 is VisionReader\textsubscript{\textit{withViT5}})}
\end{figure*}

\begin{figure*}[!h]
  \centering
  \includegraphics[width=0.9\textwidth, height=9cm]{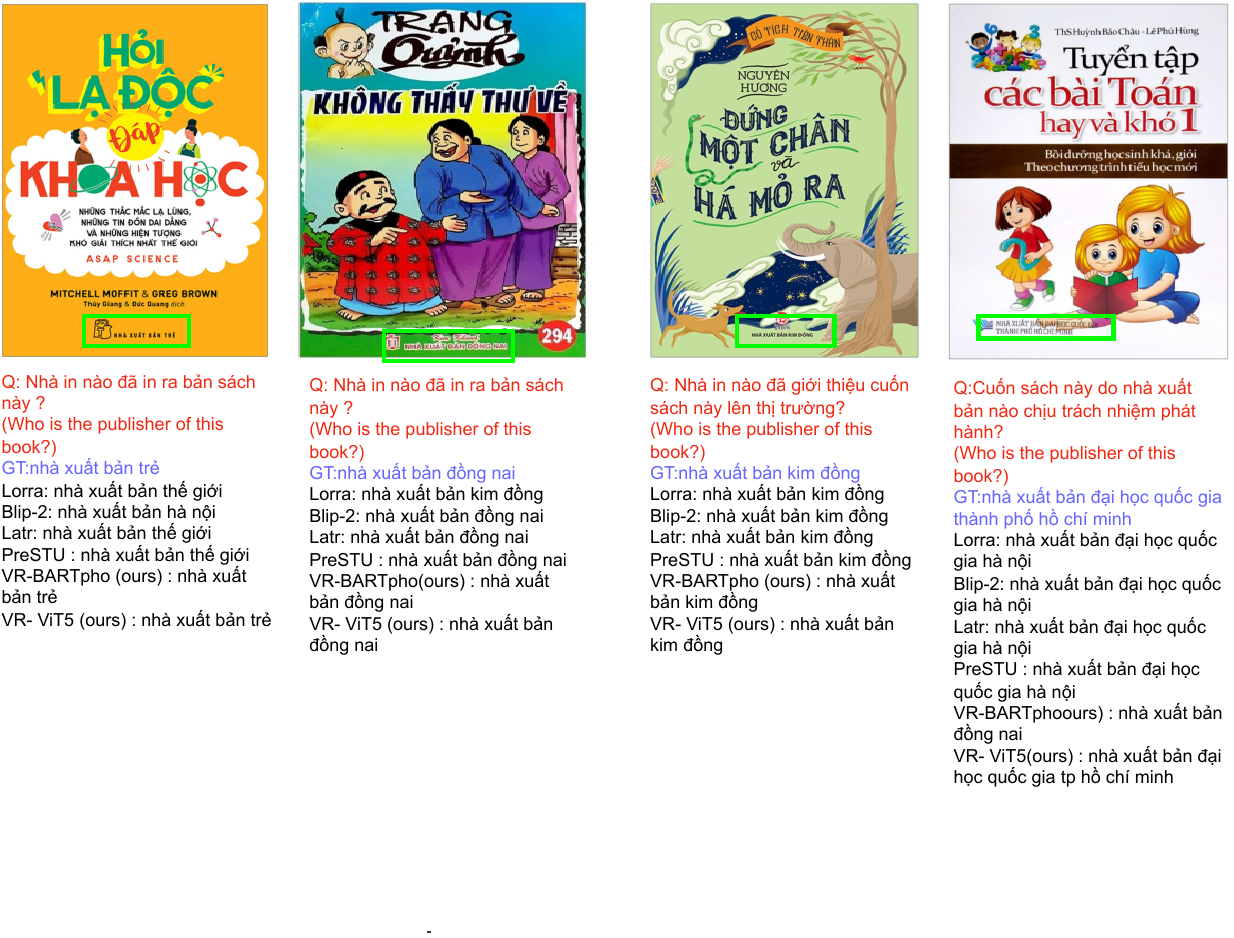}
  \caption{Examples results of models in publisher field. (Note: VR-BARTpho is VisionReader\textsubscript{\textit{withBARTpho}}, VR-ViT5 is VisionReader\textsubscript{\textit{withViT5}})}
\end{figure*}
\end{appendices}
\clearpage
\renewcommand\refname{References}
\bibliography{reference}


\begin{thebibliography}{67}
\ifx \bisbn   \undefined \def \bisbn  #1{ISBN #1}\fi
\ifx \binits  \undefined \def \binits#1{#1}\fi
\ifx \bauthor  \undefined \def \bauthor#1{#1}\fi
\ifx \batitle  \undefined \def \batitle#1{#1}\fi
\ifx \bjtitle  \undefined \def \bjtitle#1{#1}\fi
\ifx \bvolume  \undefined \def \bvolume#1{\textbf{#1}}\fi
\ifx \byear  \undefined \def \byear#1{#1}\fi
\ifx \bissue  \undefined \def \bissue#1{#1}\fi
\ifx \bfpage  \undefined \def \bfpage#1{#1}\fi
\ifx \blpage  \undefined \def \blpage #1{#1}\fi
\ifx \burl  \undefined \def \burl#1{\textsf{#1}}\fi
\ifx \doiurl  \undefined \def \doiurl#1{\url{https://doi.org/#1}}\fi
\ifx \betal  \undefined \def \betal{\textit{et al.}}\fi
\ifx \binstitute  \undefined \def \binstitute#1{#1}\fi
\ifx \binstitutionaled  \undefined \def \binstitutionaled#1{#1}\fi
\ifx \bctitle  \undefined \def \bctitle#1{#1}\fi
\ifx \beditor  \undefined \def \beditor#1{#1}\fi
\ifx \bpublisher  \undefined \def \bpublisher#1{#1}\fi
\ifx \bbtitle  \undefined \def \bbtitle#1{#1}\fi
\ifx \bedition  \undefined \def \bedition#1{#1}\fi
\ifx \bseriesno  \undefined \def \bseriesno#1{#1}\fi
\ifx \blocation  \undefined \def \blocation#1{#1}\fi
\ifx \bsertitle  \undefined \def \bsertitle#1{#1}\fi
\ifx \bsnm \undefined \def \bsnm#1{#1}\fi
\ifx \bsuffix \undefined \def \bsuffix#1{#1}\fi
\ifx \bparticle \undefined \def \bparticle#1{#1}\fi
\ifx \barticle \undefined \def \barticle#1{#1}\fi
\bibcommenthead
\ifx \bconfdate \undefined \def \bconfdate #1{#1}\fi
\ifx \botherref \undefined \def \botherref #1{#1}\fi
\ifx \url \undefined \def \url#1{\textsf{#1}}\fi
\ifx \bchapter \undefined \def \bchapter#1{#1}\fi
\ifx \bbook \undefined \def \bbook#1{#1}\fi
\ifx \bcomment \undefined \def \bcomment#1{#1}\fi
\ifx \oauthor \undefined \def \oauthor#1{#1}\fi
\ifx \citeauthoryear \undefined \def \citeauthoryear#1{#1}\fi
\ifx \endbibitem  \undefined \def \endbibitem {}\fi
\ifx \bconflocation  \undefined \def \bconflocation#1{#1}\fi
\ifx \arxivurl  \undefined \def \arxivurl#1{\textsf{#1}}\fi
\csname PreBibitemsHook\endcsname

\bibitem[\protect\citeauthoryear{Antol et~al.}{2015}]{antol2015vqa}
\begin{bchapter}
\bauthor{\bsnm{Antol}, \binits{S.}},
\bauthor{\bsnm{Agrawal}, \binits{A.}},
\bauthor{\bsnm{Lu}, \binits{J.}},
\bauthor{\bsnm{Mitchell}, \binits{M.}},
\bauthor{\bsnm{Batra}, \binits{D.}},
\bauthor{\bsnm{Zitnick}, \binits{C.L.}},
\bauthor{\bsnm{Parikh}, \binits{D.}}:
\bctitle{Vqa: Visual question answering}.
In: \bbtitle{Proceedings of the IEEE International Conference on Computer Vision},
pp. \bfpage{2425}--\blpage{2433}
(\byear{2015})
\end{bchapter}
\endbibitem

\bibitem[\protect\citeauthoryear{Goyal et~al.}{2017}]{goyal2017making}
\begin{bchapter}
\bauthor{\bsnm{Goyal}, \binits{Y.}},
\bauthor{\bsnm{Khot}, \binits{T.}},
\bauthor{\bsnm{Summers-Stay}, \binits{D.}},
\bauthor{\bsnm{Batra}, \binits{D.}},
\bauthor{\bsnm{Parikh}, \binits{D.}}:
\bctitle{Making the v in vqa matter: Elevating the role of image understanding in visual question answering}.
In: \bbtitle{Proceedings of the IEEE Conference on Computer Vision and Pattern Recognition},
pp. \bfpage{6904}--\blpage{6913}
(\byear{2017})
\end{bchapter}
\endbibitem

\bibitem[\protect\citeauthoryear{Singh et~al.}{2019}]{singh2019towards}
\begin{bchapter}
\bauthor{\bsnm{Singh}, \binits{A.}},
\bauthor{\bsnm{Natarajan}, \binits{V.}},
\bauthor{\bsnm{Shah}, \binits{M.}},
\bauthor{\bsnm{Jiang}, \binits{Y.}},
\bauthor{\bsnm{Chen}, \binits{X.}},
\bauthor{\bsnm{Batra}, \binits{D.}},
\bauthor{\bsnm{Parikh}, \binits{D.}},
\bauthor{\bsnm{Rohrbach}, \binits{M.}}:
\bctitle{Towards vqa models that can read}.
In: \bbtitle{Proceedings of the IEEE/CVF Conference on Computer Vision and Pattern Recognition},
pp. \bfpage{8317}--\blpage{8326}
(\byear{2019})
\end{bchapter}
\endbibitem

\bibitem[\protect\citeauthoryear{Biten et~al.}{2019}]{biten2019scene}
\begin{bchapter}
\bauthor{\bsnm{Biten}, \binits{A.F.}},
\bauthor{\bsnm{Tito}, \binits{R.}},
\bauthor{\bsnm{Mafla}, \binits{A.}},
\bauthor{\bsnm{Gomez}, \binits{L.}},
\bauthor{\bsnm{Rusinol}, \binits{M.}},
\bauthor{\bsnm{Valveny}, \binits{E.}},
\bauthor{\bsnm{Jawahar}, \binits{C.}},
\bauthor{\bsnm{Karatzas}, \binits{D.}}:
\bctitle{Scene text visual question answering}.
In: \bbtitle{Proceedings of the IEEE/CVF International Conference on Computer Vision},
pp. \bfpage{4291}--\blpage{4301}
(\byear{2019})
\end{bchapter}
\endbibitem

\bibitem[\protect\citeauthoryear{Kamel et~al.}{2023}]{kamel2023vaqa}
\begin{barticle}
\bauthor{\bsnm{Kamel}, \binits{S.M.}},
\bauthor{\bsnm{Hassan}, \binits{S.I.}},
\bauthor{\bsnm{Elrefaei}, \binits{L.}}:
\batitle{Vaqa: Visual arabic question answering}.
\bjtitle{Arabian Journal for Science and engineering}
\bvolume{48}(\bissue{8}),
\bfpage{10803}--\blpage{10823}
(\byear{2023})
\end{barticle}
\endbibitem

\bibitem[\protect\citeauthoryear{Kim et~al.}{2024}]{kim2024bok}
\begin{botherref}
\oauthor{\bsnm{Kim}, \binits{M.}},
\oauthor{\bsnm{Song}, \binits{S.}},
\oauthor{\bsnm{Lee}, \binits{Y.}},
\oauthor{\bsnm{Jang}, \binits{H.}},
\oauthor{\bsnm{Lim}, \binits{K.}}:
Bok-vqa: Bilingual outside knowledge-based visual question answering via graph representation pretraining.
arXiv preprint arXiv:2401.06443
(2024)
\end{botherref}
\endbibitem

\bibitem[\protect\citeauthoryear{Shimizu et~al.}{2018}]{shimizu-etal-2018-visual}
\begin{bchapter}
\bauthor{\bsnm{Shimizu}, \binits{N.}},
\bauthor{\bsnm{Rong}, \binits{N.}},
\bauthor{\bsnm{Miyazaki}, \binits{T.}}:
\bctitle{Visual question answering dataset for bilingual image understanding: A study of cross-lingual transfer using attention maps}.
In: \beditor{\bsnm{Bender}, \binits{E.M.}},
\beditor{\bsnm{Derczynski}, \binits{L.}},
\beditor{\bsnm{Isabelle}, \binits{P.}} (eds.)
\bbtitle{Proceedings of the 27th International Conference on Computational Linguistics},
pp. \bfpage{1918}--\blpage{1928}.
\bpublisher{Association for Computational Linguistics},
\blocation{Santa Fe, New Mexico, USA}
(\byear{2018})
\end{bchapter}
\endbibitem

\bibitem[\protect\citeauthoryear{Tran et~al.}{2021}]{tran2021vivqa}
\begin{bchapter}
\bauthor{\bsnm{Tran}, \binits{K.Q.}},
\bauthor{\bsnm{Nguyen}, \binits{A.T.}},
\bauthor{\bsnm{Le}, \binits{A.T.-H.}},
\bauthor{\bsnm{Van~Nguyen}, \binits{K.}}:
\bctitle{Vivqa: Vietnamese visual question answering}.
In: \bbtitle{Proceedings of the 35th Pacific Asia Conference on Language, Information and Computation},
pp. \bfpage{683}--\blpage{691}
(\byear{2021})
\end{bchapter}
\endbibitem

\bibitem[\protect\citeauthoryear{Nguyen et~al.}{2023a}]{nguyen2023openvivqa}
\begin{barticle}
\bauthor{\bsnm{Nguyen}, \binits{N.H.}},
\bauthor{\bsnm{Vo}, \binits{D.T.}},
\bauthor{\bsnm{Van~Nguyen}, \binits{K.}},
\bauthor{\bsnm{Nguyen}, \binits{N.L.-T.}}:
\batitle{Openvivqa: Task, dataset, and multimodal fusion models for visual question answering in vietnamese}.
\bjtitle{Information Fusion}
\bvolume{100},
\bfpage{101868}
(\byear{2023})
\end{barticle}
\endbibitem

\bibitem[\protect\citeauthoryear{Nguyen et~al.}{2023b}]{nguyen2023evjvqa}
\begin{botherref}
\oauthor{\bsnm{Nguyen}, \binits{N.L.-T.}},
\oauthor{\bsnm{Nguyen}, \binits{N.H.}},
\oauthor{\bsnm{Vo}, \binits{D.T.}},
\oauthor{\bsnm{Tran}, \binits{K.Q.}},
\oauthor{\bsnm{Van~Nguyen}, \binits{K.}}:
Evjvqa challenge: Multilingual visual question answering.
Journal of Computer Science and Cybernetics,
237--258
(2023)
\end{botherref}
\endbibitem

\bibitem[\protect\citeauthoryear{Tran et~al.}{2023}]{tran2023viclevr}
\begin{botherref}
\oauthor{\bsnm{Tran}, \binits{K.V.}},
\oauthor{\bsnm{Phan}, \binits{H.P.}},
\oauthor{\bsnm{Van~Nguyen}, \binits{K.}},
\oauthor{\bsnm{Nguyen}, \binits{N.L.T.}}:
Viclevr: A visual reasoning dataset and hybrid multimodal fusion model for visual question answering in vietnamese.
arXiv preprint arXiv:2310.18046
(2023)
\end{botherref}
\endbibitem

\bibitem[\protect\citeauthoryear{Lin et~al.}{2014}]{lin2014microsoft}
\begin{bchapter}
\bauthor{\bsnm{Lin}, \binits{T.-Y.}},
\bauthor{\bsnm{Maire}, \binits{M.}},
\bauthor{\bsnm{Belongie}, \binits{S.}},
\bauthor{\bsnm{Hays}, \binits{J.}},
\bauthor{\bsnm{Perona}, \binits{P.}},
\bauthor{\bsnm{Ramanan}, \binits{D.}},
\bauthor{\bsnm{Doll{\'a}r}, \binits{P.}},
\bauthor{\bsnm{Zitnick}, \binits{C.L.}}:
\bctitle{Microsoft coco: Common objects in context}.
In: \bbtitle{Computer Vision--ECCV 2014: 13th European Conference, Zurich, Switzerland, September 6-12, 2014, Proceedings, Part V 13},
pp. \bfpage{740}--\blpage{755}
(\byear{2014}).
\bcomment{Springer}
\end{bchapter}
\endbibitem

\bibitem[\protect\citeauthoryear{Kazemi and Elqursh}{2017}]{kazemi2017show}
\begin{botherref}
\oauthor{\bsnm{Kazemi}, \binits{V.}},
\oauthor{\bsnm{Elqursh}, \binits{A.}}:
Show, ask, attend, and answer: A strong baseline for visual question answering.
arXiv preprint arXiv:1704.03162
(2017)
\end{botherref}
\endbibitem

\bibitem[\protect\citeauthoryear{Lu et~al.}{2016}]{lu2016hierarchical}
\begin{botherref}
\oauthor{\bsnm{Lu}, \binits{J.}},
\oauthor{\bsnm{Yang}, \binits{J.}},
\oauthor{\bsnm{Batra}, \binits{D.}},
\oauthor{\bsnm{Parikh}, \binits{D.}}:
Hierarchical question-image co-attention for visual question answering.
Advances in neural information processing systems
\textbf{29}
(2016)
\end{botherref}
\endbibitem

\bibitem[\protect\citeauthoryear{Kim et~al.}{2016}]{kim2016multimodal}
\begin{botherref}
\oauthor{\bsnm{Kim}, \binits{J.-H.}},
\oauthor{\bsnm{Lee}, \binits{S.-W.}},
\oauthor{\bsnm{Kwak}, \binits{D.}},
\oauthor{\bsnm{Heo}, \binits{M.-O.}},
\oauthor{\bsnm{Kim}, \binits{J.}},
\oauthor{\bsnm{Ha}, \binits{J.-W.}},
\oauthor{\bsnm{Zhang}, \binits{B.-T.}}:
Multimodal residual learning for visual qa.
Advances in neural information processing systems
\textbf{29}
(2016)
\end{botherref}
\endbibitem

\bibitem[\protect\citeauthoryear{Teney et~al.}{2018}]{teney2018tips}
\begin{bchapter}
\bauthor{\bsnm{Teney}, \binits{D.}},
\bauthor{\bsnm{Anderson}, \binits{P.}},
\bauthor{\bsnm{He}, \binits{X.}},
\bauthor{\bsnm{Van Den~Hengel}, \binits{A.}}:
\bctitle{Tips and tricks for visual question answering: Learnings from the 2017 challenge}.
In: \bbtitle{Proceedings of the IEEE Conference on Computer Vision and Pattern Recognition},
pp. \bfpage{4223}--\blpage{4232}
(\byear{2018})
\end{bchapter}
\endbibitem

\bibitem[\protect\citeauthoryear{Mishra et~al.}{2019}]{mishra2019ocr}
\begin{bchapter}
\bauthor{\bsnm{Mishra}, \binits{A.}},
\bauthor{\bsnm{Shekhar}, \binits{S.}},
\bauthor{\bsnm{Singh}, \binits{A.K.}},
\bauthor{\bsnm{Chakraborty}, \binits{A.}}:
\bctitle{Ocr-vqa: Visual question answering by reading text in images}.
In: \bbtitle{2019 International Conference on Document Analysis and Recognition (ICDAR)},
pp. \bfpage{947}--\blpage{952}
(\byear{2019}).
\bcomment{IEEE}
\end{bchapter}
\endbibitem

\bibitem[\protect\citeauthoryear{Mathew et~al.}{2021}]{mathew2021docvqa}
\begin{bchapter}
\bauthor{\bsnm{Mathew}, \binits{M.}},
\bauthor{\bsnm{Karatzas}, \binits{D.}},
\bauthor{\bsnm{Jawahar}, \binits{C.}}:
\bctitle{Docvqa: A dataset for vqa on document images}.
In: \bbtitle{Proceedings of the IEEE/CVF Winter Conference on Applications of Computer Vision},
pp. \bfpage{2200}--\blpage{2209}
(\byear{2021})
\end{bchapter}
\endbibitem

\bibitem[\protect\citeauthoryear{Kantharaj et~al.}{2022}]{kantharaj2022opencqa}
\begin{bchapter}
\bauthor{\bsnm{Kantharaj}, \binits{S.}},
\bauthor{\bsnm{Do}, \binits{X.L.}},
\bauthor{\bsnm{Leong}, \binits{R.T.}},
\bauthor{\bsnm{Tan}, \binits{J.Q.}},
\bauthor{\bsnm{Hoque}, \binits{E.}},
\bauthor{\bsnm{Joty}, \binits{S.}}:
\bctitle{Opencqa: Open-ended question answering with charts}.
In: \bbtitle{Proceedings of the 2022 Conference on Empirical Methods in Natural Language Processing},
pp. \bfpage{11817}--\blpage{11837}
(\byear{2022})
\end{bchapter}
\endbibitem

\bibitem[\protect\citeauthoryear{Tanaka et~al.}{2021}]{tanaka2021visualmrc}
\begin{bchapter}
\bauthor{\bsnm{Tanaka}, \binits{R.}},
\bauthor{\bsnm{Nishida}, \binits{K.}},
\bauthor{\bsnm{Yoshida}, \binits{S.}}:
\bctitle{Visualmrc: Machine reading comprehension on document images}.
In: \bbtitle{Proceedings of the AAAI Conference on Artificial Intelligence},
vol. \bseriesno{35},
pp. \bfpage{13878}--\blpage{13888}
(\byear{2021})
\end{bchapter}
\endbibitem

\bibitem[\protect\citeauthoryear{Mathew et~al.}{2022}]{mathew2022infographicvqa}
\begin{bchapter}
\bauthor{\bsnm{Mathew}, \binits{M.}},
\bauthor{\bsnm{Bagal}, \binits{V.}},
\bauthor{\bsnm{Tito}, \binits{R.}},
\bauthor{\bsnm{Karatzas}, \binits{D.}},
\bauthor{\bsnm{Valveny}, \binits{E.}},
\bauthor{\bsnm{Jawahar}, \binits{C.}}:
\bctitle{Infographicvqa}.
In: \bbtitle{Proceedings of the IEEE/CVF Winter Conference on Applications of Computer Vision},
pp. \bfpage{1697}--\blpage{1706}
(\byear{2022})
\end{bchapter}
\endbibitem

\bibitem[\protect\citeauthoryear{Gurari et~al.}{2018}]{gurari2018vizwiz}
\begin{bchapter}
\bauthor{\bsnm{Gurari}, \binits{D.}},
\bauthor{\bsnm{Li}, \binits{Q.}},
\bauthor{\bsnm{Stangl}, \binits{A.J.}},
\bauthor{\bsnm{Guo}, \binits{A.}},
\bauthor{\bsnm{Lin}, \binits{C.}},
\bauthor{\bsnm{Grauman}, \binits{K.}},
\bauthor{\bsnm{Luo}, \binits{J.}},
\bauthor{\bsnm{Bigham}, \binits{J.P.}}:
\bctitle{Vizwiz grand challenge: Answering visual questions from blind people}.
In: \bbtitle{Proceedings of the IEEE Conference on Computer Vision and Pattern Recognition},
pp. \bfpage{3608}--\blpage{3617}
(\byear{2018})
\end{bchapter}
\endbibitem

\bibitem[\protect\citeauthoryear{Marino et~al.}{2019}]{marino2019ok}
\begin{bchapter}
\bauthor{\bsnm{Marino}, \binits{K.}},
\bauthor{\bsnm{Rastegari}, \binits{M.}},
\bauthor{\bsnm{Farhadi}, \binits{A.}},
\bauthor{\bsnm{Mottaghi}, \binits{R.}}:
\bctitle{Ok-vqa: A visual question answering benchmark requiring external knowledge}.
In: \bbtitle{Proceedings of the IEEE/cvf Conference on Computer Vision and Pattern Recognition},
pp. \bfpage{3195}--\blpage{3204}
(\byear{2019})
\end{bchapter}
\endbibitem

\bibitem[\protect\citeauthoryear{Hudson and Manning}{2019}]{hudson2019gqa}
\begin{bchapter}
\bauthor{\bsnm{Hudson}, \binits{D.A.}},
\bauthor{\bsnm{Manning}, \binits{C.D.}}:
\bctitle{Gqa: A new dataset for real-world visual reasoning and compositional question answering}.
In: \bbtitle{Proceedings of the IEEE/CVF Conference on Computer Vision and Pattern Recognition},
pp. \bfpage{6700}--\blpage{6709}
(\byear{2019})
\end{bchapter}
\endbibitem

\bibitem[\protect\citeauthoryear{Krishna et~al.}{2017}]{krishna2017visual}
\begin{barticle}
\bauthor{\bsnm{Krishna}, \binits{R.}},
\bauthor{\bsnm{Zhu}, \binits{Y.}},
\bauthor{\bsnm{Groth}, \binits{O.}},
\bauthor{\bsnm{Johnson}, \binits{J.}},
\bauthor{\bsnm{Hata}, \binits{K.}},
\bauthor{\bsnm{Kravitz}, \binits{J.}},
\bauthor{\bsnm{Chen}, \binits{S.}},
\bauthor{\bsnm{Kalantidis}, \binits{Y.}},
\bauthor{\bsnm{Li}, \binits{L.-J.}},
\bauthor{\bsnm{Shamma}, \binits{D.A.}}, \betal:
\batitle{Visual genome: Connecting language and vision using crowdsourced dense image annotations}.
\bjtitle{International journal of computer vision}
\bvolume{123},
\bfpage{32}--\blpage{73}
(\byear{2017})
\end{barticle}
\endbibitem

\bibitem[\protect\citeauthoryear{Johnson et~al.}{2017}]{johnson2017clevr}
\begin{bchapter}
\bauthor{\bsnm{Johnson}, \binits{J.}},
\bauthor{\bsnm{Hariharan}, \binits{B.}},
\bauthor{\bsnm{Van Der~Maaten}, \binits{L.}},
\bauthor{\bsnm{Fei-Fei}, \binits{L.}},
\bauthor{\bsnm{Lawrence~Zitnick}, \binits{C.}},
\bauthor{\bsnm{Girshick}, \binits{R.}}:
\bctitle{Clevr: A diagnostic dataset for compositional language and elementary visual reasoning}.
In: \bbtitle{Proceedings of the IEEE Conference on Computer Vision and Pattern Recognition},
pp. \bfpage{2901}--\blpage{2910}
(\byear{2017})
\end{bchapter}
\endbibitem

\bibitem[\protect\citeauthoryear{Hasegawa et~al.}{2023}]{hasegawa2023minecraft}
\begin{bchapter}
\bauthor{\bsnm{Hasegawa}, \binits{R.}},
\bauthor{\bsnm{Thawonmas}, \binits{R.}},
\bauthor{\bsnm{Tanabe}, \binits{J.}},
\bauthor{\bsnm{Yu}, \binits{L.}}:
\bctitle{Minecraft video aesthetics quality assessment model}.
In: \bbtitle{Proceedings of the 13th International Conference on Advances in Information Technology},
pp. \bfpage{1}--\blpage{5}
(\byear{2023})
\end{bchapter}
\endbibitem

\bibitem[\protect\citeauthoryear{Simonyan and Zisserman}{2015}]{simonyan2015very}
\begin{bchapter}
\bauthor{\bsnm{Simonyan}, \binits{K.}},
\bauthor{\bsnm{Zisserman}, \binits{A.}}:
\bctitle{Very deep convolutional networks for large-scale image recognition}.
In: \bbtitle{3rd International Conference on Learning Representations (ICLR 2015)}
(\byear{2015}).
\bcomment{Computational and Biological Learning Society}
\end{bchapter}
\endbibitem

\bibitem[\protect\citeauthoryear{Strobelt et~al.}{2017}]{strobelt2017lstmvis}
\begin{barticle}
\bauthor{\bsnm{Strobelt}, \binits{H.}},
\bauthor{\bsnm{Gehrmann}, \binits{S.}},
\bauthor{\bsnm{Pfister}, \binits{H.}},
\bauthor{\bsnm{Rush}, \binits{A.M.}}:
\batitle{Lstmvis: A tool for visual analysis of hidden state dynamics in recurrent neural networks}.
\bjtitle{IEEE transactions on visualization and computer graphics}
\bvolume{24}(\bissue{1}),
\bfpage{667}--\blpage{676}
(\byear{2017})
\end{barticle}
\endbibitem

\bibitem[\protect\citeauthoryear{Chen et~al.}{2022}]{chen2022quantum}
\begin{bchapter}
\bauthor{\bsnm{Chen}, \binits{S.Y.-C.}},
\bauthor{\bsnm{Yoo}, \binits{S.}},
\bauthor{\bsnm{Fang}, \binits{Y.-L.L.}}:
\bctitle{Quantum long short-term memory}.
In: \bbtitle{ICASSP 2022-2022 IEEE International Conference on Acoustics, Speech and Signal Processing (ICASSP)},
pp. \bfpage{8622}--\blpage{8626}
(\byear{2022}).
\bcomment{IEEE}
\end{bchapter}
\endbibitem

\bibitem[\protect\citeauthoryear{Chen et~al.}{2015}]{chen2015abc}
\begin{botherref}
\oauthor{\bsnm{Chen}, \binits{K.}},
\oauthor{\bsnm{Wang}, \binits{J.}},
\oauthor{\bsnm{Chen}, \binits{L.-C.}},
\oauthor{\bsnm{Gao}, \binits{H.}},
\oauthor{\bsnm{Xu}, \binits{W.}},
\oauthor{\bsnm{Nevatia}, \binits{R.}}:
Abc-cnn: An attention based convolutional neural network for visual question answering.
arXiv preprint arXiv:1511.05960
(2015)
\end{botherref}
\endbibitem

\bibitem[\protect\citeauthoryear{Ma et~al.}{2016}]{ma2016learning}
\begin{bchapter}
\bauthor{\bsnm{Ma}, \binits{L.}},
\bauthor{\bsnm{Lu}, \binits{Z.}},
\bauthor{\bsnm{Li}, \binits{H.}}:
\bctitle{Learning to answer questions from image using convolutional neural network}.
In: \bbtitle{Proceedings of the AAAI Conference on Artificial Intelligence},
vol. \bseriesno{30}
(\byear{2016})
\end{bchapter}
\endbibitem

\bibitem[\protect\citeauthoryear{Zhu et~al.}{2016}]{zhu2016visual7w}
\begin{bchapter}
\bauthor{\bsnm{Zhu}, \binits{Y.}},
\bauthor{\bsnm{Groth}, \binits{O.}},
\bauthor{\bsnm{Bernstein}, \binits{M.}},
\bauthor{\bsnm{Fei-Fei}, \binits{L.}}:
\bctitle{Visual7w: Grounded question answering in images}.
In: \bbtitle{Proceedings of the IEEE Conference on Computer Vision and Pattern Recognition},
pp. \bfpage{4995}--\blpage{5004}
(\byear{2016})
\end{bchapter}
\endbibitem

\bibitem[\protect\citeauthoryear{Shih et~al.}{2016}]{shih2016look}
\begin{bchapter}
\bauthor{\bsnm{Shih}, \binits{K.J.}},
\bauthor{\bsnm{Singh}, \binits{S.}},
\bauthor{\bsnm{Hoiem}, \binits{D.}}:
\bctitle{Where to look: Focus regions for visual question answering}.
In: \bbtitle{Proceedings of the IEEE Conference on Computer Vision and Pattern Recognition},
pp. \bfpage{4613}--\blpage{4621}
(\byear{2016})
\end{bchapter}
\endbibitem

\bibitem[\protect\citeauthoryear{Wu et~al.}{2017}]{wu2017image}
\begin{barticle}
\bauthor{\bsnm{Wu}, \binits{Q.}},
\bauthor{\bsnm{Shen}, \binits{C.}},
\bauthor{\bsnm{Wang}, \binits{P.}},
\bauthor{\bsnm{Dick}, \binits{A.}},
\bauthor{\bsnm{Van Den~Hengel}, \binits{A.}}:
\batitle{Image captioning and visual question answering based on attributes and external knowledge}.
\bjtitle{IEEE transactions on pattern analysis and machine intelligence}
\bvolume{40}(\bissue{6}),
\bfpage{1367}--\blpage{1381}
(\byear{2017})
\end{barticle}
\endbibitem

\bibitem[\protect\citeauthoryear{Kenton and Toutanova}{2019}]{kenton2019bert}
\begin{bchapter}
\bauthor{\bsnm{Kenton}, \binits{J.D.M.-W.C.}},
\bauthor{\bsnm{Toutanova}, \binits{L.K.}}:
\bctitle{Bert: Pre-training of deep bidirectional transformers for language understanding}.
In: \bbtitle{Proceedings of NAACL-HLT},
pp. \bfpage{4171}--\blpage{4186}
(\byear{2019})
\end{bchapter}
\endbibitem

\bibitem[\protect\citeauthoryear{Lu et~al.}{2019}]{lu2019vilbert}
\begin{botherref}
\oauthor{\bsnm{Lu}, \binits{J.}},
\oauthor{\bsnm{Batra}, \binits{D.}},
\oauthor{\bsnm{Parikh}, \binits{D.}},
\oauthor{\bsnm{Lee}, \binits{S.}}:
Vilbert: Pretraining task-agnostic visiolinguistic representations for vision-and-language tasks.
Advances in neural information processing systems
\textbf{32}
(2019)
\end{botherref}
\endbibitem

\bibitem[\protect\citeauthoryear{Li et~al.}{2019}]{li2019visualbert}
\begin{botherref}
\oauthor{\bsnm{Li}, \binits{L.H.}},
\oauthor{\bsnm{Yatskar}, \binits{M.}},
\oauthor{\bsnm{Yin}, \binits{D.}},
\oauthor{\bsnm{Hsieh}, \binits{C.-J.}},
\oauthor{\bsnm{Chang}, \binits{K.-W.}}:
Visualbert: A simple and performant baseline for vision and language.
arXiv preprint arXiv:1908.03557
(2019)
\end{botherref}
\endbibitem

\bibitem[\protect\citeauthoryear{Tan and Bansal}{2019}]{tan2019lxmert}
\begin{bchapter}
\bauthor{\bsnm{Tan}, \binits{H.}},
\bauthor{\bsnm{Bansal}, \binits{M.}}:
\bctitle{Lxmert: Learning cross-modality encoder representations from transformers}.
In: \bbtitle{Proceedings of the 2019 Conference on Empirical Methods in Natural Language Processing and the 9th International Joint Conference on Natural Language Processing (EMNLP-IJCNLP)},
pp. \bfpage{5100}--\blpage{5111}
(\byear{2019})
\end{bchapter}
\endbibitem

\bibitem[\protect\citeauthoryear{Su et~al.}{2019}]{su2019vl}
\begin{bchapter}
\bauthor{\bsnm{Su}, \binits{W.}},
\bauthor{\bsnm{Zhu}, \binits{X.}},
\bauthor{\bsnm{Cao}, \binits{Y.}},
\bauthor{\bsnm{Li}, \binits{B.}},
\bauthor{\bsnm{Lu}, \binits{L.}},
\bauthor{\bsnm{Wei}, \binits{F.}},
\bauthor{\bsnm{Dai}, \binits{J.}}:
\bctitle{Vl-bert: Pre-training of generic visual-linguistic representations}.
In: \bbtitle{International Conference on Learning Representations}
(\byear{2019})
\end{bchapter}
\endbibitem

\bibitem[\protect\citeauthoryear{Chen et~al.}{2020}]{chen2020uniter}
\begin{bchapter}
\bauthor{\bsnm{Chen}, \binits{Y.-C.}},
\bauthor{\bsnm{Li}, \binits{L.}},
\bauthor{\bsnm{Yu}, \binits{L.}},
\bauthor{\bsnm{El~Kholy}, \binits{A.}},
\bauthor{\bsnm{Ahmed}, \binits{F.}},
\bauthor{\bsnm{Gan}, \binits{Z.}},
\bauthor{\bsnm{Cheng}, \binits{Y.}},
\bauthor{\bsnm{Liu}, \binits{J.}}:
\bctitle{Uniter: Universal image-text representation learning}.
In: \bbtitle{European Conference on Computer Vision},
pp. \bfpage{104}--\blpage{120}
(\byear{2020}).
\bcomment{Springer}
\end{bchapter}
\endbibitem

\bibitem[\protect\citeauthoryear{Li et~al.}{2020}]{li2020oscar}
\begin{bchapter}
\bauthor{\bsnm{Li}, \binits{X.}},
\bauthor{\bsnm{Yin}, \binits{X.}},
\bauthor{\bsnm{Li}, \binits{C.}},
\bauthor{\bsnm{Zhang}, \binits{P.}},
\bauthor{\bsnm{Hu}, \binits{X.}},
\bauthor{\bsnm{Zhang}, \binits{L.}},
\bauthor{\bsnm{Wang}, \binits{L.}},
\bauthor{\bsnm{Hu}, \binits{H.}},
\bauthor{\bsnm{Dong}, \binits{L.}},
\bauthor{\bsnm{Wei}, \binits{F.}}, \betal:
\bctitle{Oscar: Object-semantics aligned pre-training for vision-language tasks}.
In: \bbtitle{Computer Vision--ECCV 2020: 16th European Conference, Glasgow, UK, August 23--28, 2020, Proceedings, Part XXX 16},
pp. \bfpage{121}--\blpage{137}
(\byear{2020}).
\bcomment{Springer}
\end{bchapter}
\endbibitem

\bibitem[\protect\citeauthoryear{Raffel et~al.}{2020}]{raffel2020exploring}
\begin{barticle}
\bauthor{\bsnm{Raffel}, \binits{C.}},
\bauthor{\bsnm{Shazeer}, \binits{N.}},
\bauthor{\bsnm{Roberts}, \binits{A.}},
\bauthor{\bsnm{Lee}, \binits{K.}},
\bauthor{\bsnm{Narang}, \binits{S.}},
\bauthor{\bsnm{Matena}, \binits{M.}},
\bauthor{\bsnm{Zhou}, \binits{Y.}},
\bauthor{\bsnm{Li}, \binits{W.}},
\bauthor{\bsnm{Liu}, \binits{P.J.}}:
\batitle{Exploring the limits of transfer learning with a unified text-to-text transformer}.
\bjtitle{The Journal of Machine Learning Research}
\bvolume{21}(\bissue{1}),
\bfpage{5485}--\blpage{5551}
(\byear{2020})
\end{barticle}
\endbibitem

\bibitem[\protect\citeauthoryear{Biten et~al.}{2022}]{biten2022latr}
\begin{bchapter}
\bauthor{\bsnm{Biten}, \binits{A.F.}},
\bauthor{\bsnm{Litman}, \binits{R.}},
\bauthor{\bsnm{Xie}, \binits{Y.}},
\bauthor{\bsnm{Appalaraju}, \binits{S.}},
\bauthor{\bsnm{Manmatha}, \binits{R.}}:
\bctitle{Latr: Layout-aware transformer for scene-text vqa}.
In: \bbtitle{Proceedings of the IEEE/CVF Conference on Computer Vision and Pattern Recognition},
pp. \bfpage{16548}--\blpage{16558}
(\byear{2022})
\end{bchapter}
\endbibitem

\bibitem[\protect\citeauthoryear{Kil et~al.}{2023}]{kil2023prestu}
\begin{bchapter}
\bauthor{\bsnm{Kil}, \binits{J.}},
\bauthor{\bsnm{Changpinyo}, \binits{S.}},
\bauthor{\bsnm{Chen}, \binits{X.}},
\bauthor{\bsnm{Hu}, \binits{H.}},
\bauthor{\bsnm{Goodman}, \binits{S.}},
\bauthor{\bsnm{Chao}, \binits{W.-L.}},
\bauthor{\bsnm{Soricut}, \binits{R.}}:
\bctitle{Prestu: Pre-training for scene-text understanding}.
In: \bbtitle{Proceedings of the IEEE/CVF International Conference on Computer Vision},
pp. \bfpage{15270}--\blpage{15280}
(\byear{2023})
\end{bchapter}
\endbibitem

\bibitem[\protect\citeauthoryear{Cho et~al.}{2021}]{cho2021unifying}
\begin{bchapter}
\bauthor{\bsnm{Cho}, \binits{J.}},
\bauthor{\bsnm{Lei}, \binits{J.}},
\bauthor{\bsnm{Tan}, \binits{H.}},
\bauthor{\bsnm{Bansal}, \binits{M.}}:
\bctitle{Unifying vision-and-language tasks via text generation}.
In: \bbtitle{International Conference on Machine Learning},
pp. \bfpage{1931}--\blpage{1942}
(\byear{2021}).
\bcomment{PMLR}
\end{bchapter}
\endbibitem

\bibitem[\protect\citeauthoryear{Fang et~al.}{2023}]{fang2023separate}
\begin{bchapter}
\bauthor{\bsnm{Fang}, \binits{C.}},
\bauthor{\bsnm{Li}, \binits{J.}},
\bauthor{\bsnm{Li}, \binits{L.}},
\bauthor{\bsnm{Ma}, \binits{C.}},
\bauthor{\bsnm{Hu}, \binits{D.}}:
\bctitle{Separate and locate: Rethink the text in text-based visual question answering}.
In: \bbtitle{Proceedings of the 31st ACM International Conference on Multimedia},
pp. \bfpage{4378}--\blpage{4388}
(\byear{2023})
\end{bchapter}
\endbibitem

\bibitem[\protect\citeauthoryear{Li et~al.}{2023}]{li2023blip}
\begin{bchapter}
\bauthor{\bsnm{Li}, \binits{J.}},
\bauthor{\bsnm{Li}, \binits{D.}},
\bauthor{\bsnm{Savarese}, \binits{S.}},
\bauthor{\bsnm{Hoi}, \binits{S.}}:
\bctitle{Blip-2: Bootstrapping language-image pre-training with frozen image encoders and large language models}.
In: \bbtitle{International Conference on Machine Learning},
pp. \bfpage{19730}--\blpage{19742}
(\byear{2023}).
\bcomment{PMLR}
\end{bchapter}
\endbibitem

\bibitem[\protect\citeauthoryear{Alayrac et~al.}{2022}]{alayrac2022flamingo}
\begin{barticle}
\bauthor{\bsnm{Alayrac}, \binits{J.-B.}},
\bauthor{\bsnm{Donahue}, \binits{J.}},
\bauthor{\bsnm{Luc}, \binits{P.}},
\bauthor{\bsnm{Miech}, \binits{A.}},
\bauthor{\bsnm{Barr}, \binits{I.}},
\bauthor{\bsnm{Hasson}, \binits{Y.}},
\bauthor{\bsnm{Lenc}, \binits{K.}},
\bauthor{\bsnm{Mensch}, \binits{A.}},
\bauthor{\bsnm{Millican}, \binits{K.}},
\bauthor{\bsnm{Reynolds}, \binits{M.}}, \betal:
\batitle{Flamingo: a visual language model for few-shot learning}.
\bjtitle{Advances in Neural Information Processing Systems}
\bvolume{35},
\bfpage{23716}--\blpage{23736}
(\byear{2022})
\end{barticle}
\endbibitem

\bibitem[\protect\citeauthoryear{Singh et~al.}{2022}]{singh2022flava}
\begin{bchapter}
\bauthor{\bsnm{Singh}, \binits{A.}},
\bauthor{\bsnm{Hu}, \binits{R.}},
\bauthor{\bsnm{Goswami}, \binits{V.}},
\bauthor{\bsnm{Couairon}, \binits{G.}},
\bauthor{\bsnm{Galuba}, \binits{W.}},
\bauthor{\bsnm{Rohrbach}, \binits{M.}},
\bauthor{\bsnm{Kiela}, \binits{D.}}:
\bctitle{Flava: A foundational language and vision alignment model}.
In: \bbtitle{Proceedings of the IEEE/CVF Conference on Computer Vision and Pattern Recognition},
pp. \bfpage{15638}--\blpage{15650}
(\byear{2022})
\end{bchapter}
\endbibitem

\bibitem[\protect\citeauthoryear{Li et~al.}{2022}]{li2022mplug}
\begin{bchapter}
\bauthor{\bsnm{Li}, \binits{C.}},
\bauthor{\bsnm{Xu}, \binits{H.}},
\bauthor{\bsnm{Tian}, \binits{J.}},
\bauthor{\bsnm{Wang}, \binits{W.}},
\bauthor{\bsnm{Yan}, \binits{M.}},
\bauthor{\bsnm{Bi}, \binits{B.}},
\bauthor{\bsnm{Ye}, \binits{J.}},
\bauthor{\bsnm{Chen}, \binits{H.}},
\bauthor{\bsnm{Xu}, \binits{G.}},
\bauthor{\bsnm{Cao}, \binits{Z.}}, \betal:
\bctitle{mplug: Effective and efficient vision-language learning by cross-modal skip-connections}.
In: \bbtitle{Proceedings of the 2022 Conference on Empirical Methods in Natural Language Processing},
pp. \bfpage{7241}--\blpage{7259}
(\byear{2022})
\end{bchapter}
\endbibitem

\bibitem[\protect\citeauthoryear{Maronga et~al.}{2020}]{maronga2020overview}
\begin{barticle}
\bauthor{\bsnm{Maronga}, \binits{B.}},
\bauthor{\bsnm{Banzhaf}, \binits{S.}},
\bauthor{\bsnm{Burmeister}, \binits{C.}},
\bauthor{\bsnm{Esch}, \binits{T.}},
\bauthor{\bsnm{Forkel}, \binits{R.}},
\bauthor{\bsnm{Fr{\"o}hlich}, \binits{D.}},
\bauthor{\bsnm{Fuka}, \binits{V.}},
\bauthor{\bsnm{Gehrke}, \binits{K.F.}},
\bauthor{\bsnm{Geleti{\v{c}}}, \binits{J.}},
\bauthor{\bsnm{Giersch}, \binits{S.}}, \betal:
\batitle{Overview of the palm model system 6.0}.
\bjtitle{Geoscientific Model Development}
\bvolume{13}(\bissue{3}),
\bfpage{1335}--\blpage{1372}
(\byear{2020})
\end{barticle}
\endbibitem

\bibitem[\protect\citeauthoryear{Anil et~al.}{2023}]{anil2023palm}
\begin{botherref}
\oauthor{\bsnm{Anil}, \binits{R.}},
\oauthor{\bsnm{Dai}, \binits{A.M.}},
\oauthor{\bsnm{Firat}, \binits{O.}},
\oauthor{\bsnm{Johnson}, \binits{M.}},
\oauthor{\bsnm{Lepikhin}, \binits{D.}},
\oauthor{\bsnm{Passos}, \binits{A.}},
\oauthor{\bsnm{Shakeri}, \binits{S.}},
\oauthor{\bsnm{Taropa}, \binits{E.}},
\oauthor{\bsnm{Bailey}, \binits{P.}},
\oauthor{\bsnm{Chen}, \binits{Z.}}, et al.:
Palm 2 technical report.
arXiv preprint arXiv:2305.10403
(2023)
\end{botherref}
\endbibitem

\bibitem[\protect\citeauthoryear{Phan et~al.}{2022}]{phan2022vit5}
\begin{bchapter}
\bauthor{\bsnm{Phan}, \binits{L.}},
\bauthor{\bsnm{Tran}, \binits{H.}},
\bauthor{\bsnm{Nguyen}, \binits{H.}},
\bauthor{\bsnm{Trinh}, \binits{T.H.}}:
\bctitle{Vit5: Pretrained text-to-text transformer for vietnamese language generation}.
In: \bbtitle{Proceedings of the 2022 Conference of the North American Chapter of the Association for Computational Linguistics: Human Language Technologies: Student Research Workshop},
pp. \bfpage{136}--\blpage{142}
(\byear{2022})
\end{bchapter}
\endbibitem

\bibitem[\protect\citeauthoryear{Tran et~al.}{2022}]{bartpho}
\begin{bchapter}
\bauthor{\bsnm{Tran}, \binits{N.L.}},
\bauthor{\bsnm{Le}, \binits{D.M.}},
\bauthor{\bsnm{Nguyen}, \binits{D.Q.}}:
\bctitle{{BARTpho: Pre-trained Sequence-to-Sequence Models for Vietnamese}}.
In: \bbtitle{Proceedings of the 23rd Annual Conference of the International Speech Communication Association}
(\byear{2022})
\end{bchapter}
\endbibitem

\bibitem[\protect\citeauthoryear{Lewis et~al.}{2020}]{lewis2020bart}
\begin{bchapter}
\bauthor{\bsnm{Lewis}, \binits{M.}},
\bauthor{\bsnm{Liu}, \binits{Y.}},
\bauthor{\bsnm{Goyal}, \binits{N.}},
\bauthor{\bsnm{Ghazvininejad}, \binits{M.}},
\bauthor{\bsnm{Mohamed}, \binits{A.}},
\bauthor{\bsnm{Levy}, \binits{O.}},
\bauthor{\bsnm{Stoyanov}, \binits{V.}},
\bauthor{\bsnm{Zettlemoyer}, \binits{L.}}:
\bctitle{Bart: Denoising sequence-to-sequence pre-training for natural language generation, translation, and comprehension}.
In: \bbtitle{Proceedings of the 58th Annual Meeting of the Association for Computational Linguistics}
(\byear{2020}).
\bcomment{Association for Computational Linguistics}
\end{bchapter}
\endbibitem

\bibitem[\protect\citeauthoryear{Zhang et~al.}{2021}]{zhang2021vinvl}
\begin{bchapter}
\bauthor{\bsnm{Zhang}, \binits{P.}},
\bauthor{\bsnm{Li}, \binits{X.}},
\bauthor{\bsnm{Hu}, \binits{X.}},
\bauthor{\bsnm{Yang}, \binits{J.}},
\bauthor{\bsnm{Zhang}, \binits{L.}},
\bauthor{\bsnm{Wang}, \binits{L.}},
\bauthor{\bsnm{Choi}, \binits{Y.}},
\bauthor{\bsnm{Gao}, \binits{J.}}:
\bctitle{Vinvl: Revisiting visual representations in vision-language models}.
In: \bbtitle{Proceedings of the IEEE/CVF Conference on Computer Vision and Pattern Recognition},
pp. \bfpage{5579}--\blpage{5588}
(\byear{2021})
\end{bchapter}
\endbibitem

\bibitem[\protect\citeauthoryear{Huang et~al.}{2022}]{huang2022swintextspotter}
\begin{bchapter}
\bauthor{\bsnm{Huang}, \binits{M.}},
\bauthor{\bsnm{Liu}, \binits{Y.}},
\bauthor{\bsnm{Peng}, \binits{Z.}},
\bauthor{\bsnm{Liu}, \binits{C.}},
\bauthor{\bsnm{Lin}, \binits{D.}},
\bauthor{\bsnm{Zhu}, \binits{S.}},
\bauthor{\bsnm{Yuan}, \binits{N.}},
\bauthor{\bsnm{Ding}, \binits{K.}},
\bauthor{\bsnm{Jin}, \binits{L.}}:
\bctitle{Swintextspotter: Scene text spotting via better synergy between text detection and text recognition}.
In: \bbtitle{Proceedings of the IEEE/CVF Conference on Computer Vision and Pattern Recognition},
pp. \bfpage{4593}--\blpage{4603}
(\byear{2022})
\end{bchapter}
\endbibitem

\bibitem[\protect\citeauthoryear{Dosovitskiy et~al.}{2020}]{dosovitskiy2020image}
\begin{bchapter}
\bauthor{\bsnm{Dosovitskiy}, \binits{A.}},
\bauthor{\bsnm{Beyer}, \binits{L.}},
\bauthor{\bsnm{Kolesnikov}, \binits{A.}},
\bauthor{\bsnm{Weissenborn}, \binits{D.}},
\bauthor{\bsnm{Zhai}, \binits{X.}},
\bauthor{\bsnm{Unterthiner}, \binits{T.}},
\bauthor{\bsnm{Dehghani}, \binits{M.}},
\bauthor{\bsnm{Minderer}, \binits{M.}},
\bauthor{\bsnm{Heigold}, \binits{G.}},
\bauthor{\bsnm{Gelly}, \binits{S.}}, \betal:
\bctitle{An image is worth 16x16 words: Transformers for image recognition at scale}.
In: \bbtitle{International Conference on Learning Representations}
(\byear{2020})
\end{bchapter}
\endbibitem

\bibitem[\protect\citeauthoryear{Radford et~al.}{2021}]{radford2021learning}
\begin{bchapter}
\bauthor{\bsnm{Radford}, \binits{A.}},
\bauthor{\bsnm{Kim}, \binits{J.W.}},
\bauthor{\bsnm{Hallacy}, \binits{C.}},
\bauthor{\bsnm{Ramesh}, \binits{A.}},
\bauthor{\bsnm{Goh}, \binits{G.}},
\bauthor{\bsnm{Agarwal}, \binits{S.}},
\bauthor{\bsnm{Sastry}, \binits{G.}},
\bauthor{\bsnm{Askell}, \binits{A.}},
\bauthor{\bsnm{Mishkin}, \binits{P.}},
\bauthor{\bsnm{Clark}, \binits{J.}}, \betal:
\bctitle{Learning transferable visual models from natural language supervision}.
In: \bbtitle{International Conference on Machine Learning},
pp. \bfpage{8748}--\blpage{8763}
(\byear{2021}).
\bcomment{PMLR}
\end{bchapter}
\endbibitem

\bibitem[\protect\citeauthoryear{Kingma and Ba}{2015}]{KingBa15}
\begin{bchapter}
\bauthor{\bsnm{Kingma}, \binits{D.}},
\bauthor{\bsnm{Ba}, \binits{J.}}:
\bctitle{Adam: A method for stochastic optimization}.
In: \bbtitle{International Conference on Learning Representations (ICLR)}
(\byear{2015})
\end{bchapter}
\endbibitem

\bibitem[\protect\citeauthoryear{Chen et~al.}{2022}]{chen2022visualgpt}
\begin{bchapter}
\bauthor{\bsnm{Chen}, \binits{J.}},
\bauthor{\bsnm{Guo}, \binits{H.}},
\bauthor{\bsnm{Yi}, \binits{K.}},
\bauthor{\bsnm{Li}, \binits{B.}},
\bauthor{\bsnm{Elhoseiny}, \binits{M.}}:
\bctitle{Visualgpt: Data-efficient adaptation of pretrained language models for image captioning}.
In: \bbtitle{Proceedings of the IEEE/CVF Conference on Computer Vision and Pattern Recognition},
pp. \bfpage{18030}--\blpage{18040}
(\byear{2022})
\end{bchapter}
\endbibitem

\bibitem[\protect\citeauthoryear{Wang et~al.}{2024}]{wang2024visionllm}
\begin{botherref}
\oauthor{\bsnm{Wang}, \binits{W.}},
\oauthor{\bsnm{Chen}, \binits{Z.}},
\oauthor{\bsnm{Chen}, \binits{X.}},
\oauthor{\bsnm{Wu}, \binits{J.}},
\oauthor{\bsnm{Zhu}, \binits{X.}},
\oauthor{\bsnm{Zeng}, \binits{G.}},
\oauthor{\bsnm{Luo}, \binits{P.}},
\oauthor{\bsnm{Lu}, \binits{T.}},
\oauthor{\bsnm{Zhou}, \binits{J.}},
\oauthor{\bsnm{Qiao}, \binits{Y.}}, et al.:
Visionllm: Large language model is also an open-ended decoder for vision-centric tasks.
Advances in Neural Information Processing Systems
\textbf{36}
(2024)
\end{botherref}
\endbibitem

\bibitem[\protect\citeauthoryear{Brown et~al.}{2020}]{brown2020language}
\begin{barticle}
\bauthor{\bsnm{Brown}, \binits{T.}},
\bauthor{\bsnm{Mann}, \binits{B.}},
\bauthor{\bsnm{Ryder}, \binits{N.}},
\bauthor{\bsnm{Subbiah}, \binits{M.}},
\bauthor{\bsnm{Kaplan}, \binits{J.D.}},
\bauthor{\bsnm{Dhariwal}, \binits{P.}},
\bauthor{\bsnm{Neelakantan}, \binits{A.}},
\bauthor{\bsnm{Shyam}, \binits{P.}},
\bauthor{\bsnm{Sastry}, \binits{G.}},
\bauthor{\bsnm{Askell}, \binits{A.}}, \betal:
\batitle{Language models are few-shot learners}.
\bjtitle{Advances in neural information processing systems}
\bvolume{33},
\bfpage{1877}--\blpage{1901}
(\byear{2020})
\end{barticle}
\endbibitem

\bibitem[\protect\citeauthoryear{Touvron et~al.}{2023}]{touvron2023llama}
\begin{botherref}
\oauthor{\bsnm{Touvron}, \binits{H.}},
\oauthor{\bsnm{Lavril}, \binits{T.}},
\oauthor{\bsnm{Izacard}, \binits{G.}},
\oauthor{\bsnm{Martinet}, \binits{X.}},
\oauthor{\bsnm{Lachaux}, \binits{M.-A.}},
\oauthor{\bsnm{Lacroix}, \binits{T.}},
\oauthor{\bsnm{Rozi{\`e}re}, \binits{B.}},
\oauthor{\bsnm{Goyal}, \binits{N.}},
\oauthor{\bsnm{Hambro}, \binits{E.}},
\oauthor{\bsnm{Azhar}, \binits{F.}}, et al.:
Llama: Open and efficient foundation language models.
arXiv preprint arXiv:2302.13971
(2023)
\end{botherref}
\endbibitem

\bibitem[\protect\citeauthoryear{Liu et~al.}{2018}]{liu2018fots}
\begin{bchapter}
\bauthor{\bsnm{Liu}, \binits{X.}},
\bauthor{\bsnm{Liang}, \binits{D.}},
\bauthor{\bsnm{Yan}, \binits{S.}},
\bauthor{\bsnm{Chen}, \binits{D.}},
\bauthor{\bsnm{Qiao}, \binits{Y.}},
\bauthor{\bsnm{Yan}, \binits{J.}}:
\bctitle{Fots: Fast oriented text spotting with a unified network}.
In: \bbtitle{Proceedings of the IEEE Conference on Computer Vision and Pattern Recognition},
pp. \bfpage{5676}--\blpage{5685}
(\byear{2018})
\end{bchapter}
\endbibitem

\bibitem[\protect\citeauthoryear{Lyu et~al.}{2018}]{lyu2018mask}
\begin{bchapter}
\bauthor{\bsnm{Lyu}, \binits{P.}},
\bauthor{\bsnm{Liao}, \binits{M.}},
\bauthor{\bsnm{Yao}, \binits{C.}},
\bauthor{\bsnm{Wu}, \binits{W.}},
\bauthor{\bsnm{Bai}, \binits{X.}}:
\bctitle{Mask textspotter: An end-to-end trainable neural network for spotting text with arbitrary shapes}.
In: \bbtitle{Proceedings of the European Conference on Computer Vision (ECCV)},
pp. \bfpage{67}--\blpage{83}
(\byear{2018})
\end{bchapter}
\endbibitem

\end{thebibliography}

\end{document}